\documentclass[11pt]{article}
\usepackage{booktabs}
\usepackage{subcaption}

\PassOptionsToPackage{hyperfootnotes=false}{hyperref}
\usepackage[final]{acl}

\usepackage{times}
\usepackage{latexsym}
\usepackage{amsmath}
\usepackage{amssymb}

\usepackage[T1]{fontenc}

\usepackage[utf8]{inputenc}

\usepackage{microtype}
\usepackage{xspace}
\usepackage{algorithm}
\usepackage{algorithmicx}
\usepackage{algpseudocode}

\usepackage{inconsolata}

\usepackage{graphicx}
\usepackage{fontawesome5}
\usepackage{array}
\usepackage{tabularx}

\usepackage{listings}
\usepackage{needspace}
\usepackage[table]{xcolor}
\definecolor{lightblue}{RGB}{235,242,249}
\definecolor{lightred}{RGB}{250,230,235}
\definecolor{jpblue}{RGB}{33, 99, 154}
\definecolor{jpred}{RGB}{188, 0, 45}

\lstdefinestyle{sqlstyle}{
    basicstyle=\ttfamily\small,
    breaklines=true,
    breakatwhitespace=false,
    columns=fullflexible,
    keepspaces=true,
    frame=single,
    rulecolor=\color{black!20},
    xleftmargin=2pt,
    xrightmargin=2pt,
    aboveskip=4pt,
    belowskip=4pt
}

\definecolor{tableLavender}{HTML}{B0B0E3}
\definecolor{tableLilac}{HTML}{E0E0F4}
\definecolor{tableBlue}{HTML}{A0BEE9}
\definecolor{tableHeader}{HTML}{C9DAF3}
\definecolor{tableCyan}{HTML}{BFE3EE}
\definecolor{tableMint}{HTML}{D6EFF6}
\colorlet{modelGPT}{tableLilac!65!white}
\colorlet{modelClaude}{tableLavender!22!white}
\colorlet{modelDeepSeek}{tableBlue!27!white}
\colorlet{modelGemini}{tableCyan!40!white}
\newcommand{\modelname}[1]{\csname modelname@#1\endcsname}
\newcommand{\modelentry}[1]{\mbox{\raisebox{-0.12em}{\includegraphics[width=0.95em,height=0.95em,keepaspectratio]{img/model-icons/\csname modelicon@#1\endcsname.pdf}}\hspace{0.3em}\modelname{#1}}}
\expandafter\def\csname modelname@gpt5mini\endcsname{GPT-5-mini}
\expandafter\def\csname modelicon@gpt5mini\endcsname{openai}
\expandafter\def\csname modelname@gpt41mini\endcsname{GPT-4.1-mini}
\expandafter\def\csname modelicon@gpt41mini\endcsname{openai}
\expandafter\def\csname modelname@claudesonnet\endcsname{Claude-4.5-Sonnet}
\expandafter\def\csname modelicon@claudesonnet\endcsname{claude}
\expandafter\def\csname modelname@claudeopus\endcsname{Claude-4.5-Opus}
\expandafter\def\csname modelicon@claudeopus\endcsname{claude}
\expandafter\def\csname modelname@deepseek\endcsname{DeepSeek-V3.1}
\expandafter\def\csname modelicon@deepseek\endcsname{deepseek}
\expandafter\def\csname modelname@geminiflash\endcsname{Gemini-2.5-Flash}
\expandafter\def\csname modelicon@geminiflash\endcsname{gemini}
\expandafter\def\csname modelname@geminipro\endcsname{Gemini-3-Pro}
\expandafter\def\csname modelicon@geminipro\endcsname{gemini}

\newcommand{\method}{\textbf{SQLStructEval}\xspace}
\newcommand{\problem}{\textit{ECSD}\xspace}
\title{\method: Structural Evaluation of \\ LLM Text-to-SQL Generation}
\author{
\textbf{Yixi Zhou}\textsuperscript{1,2}\thanks{Equal contribution.}\quad
\textbf{Fan Zhang}\textsuperscript{3}\textsuperscript{*}\quad
\textbf{Zhiqiao Guo}\textsuperscript{2,4}\textsuperscript{*}\quad
\textbf{Yu Chen}\textsuperscript{3}\thanks{Corresponding authors.}\\
\textbf{Haipeng Zhang}\textsuperscript{2}\textsuperscript{\ensuremath{\dagger}}\quad
\textbf{Preslav Nakov}\textsuperscript{5}\quad
\textbf{Zhuohan Xie}\textsuperscript{5}\\
\textsuperscript{1}Hong Kong Baptist University\quad
\textsuperscript{2}ShanghaiTech University\\
\textsuperscript{3}The University of Tokyo\quad
\textsuperscript{4}University of Michigan\quad
\textsuperscript{5}MBZUAI\\
\texttt{yxzhou@comp.hkbu.edu.hk}, \texttt{zhang-fan@g.ecc.u-tokyo.ac.jp}\\
\texttt{guozhq@umich.edu}, \texttt{chen@edu.k.u-tokyo.ac.jp}, \texttt{zhanghp@shanghaitech.edu.cn}\\
\texttt{\{preslav.nakov,zhuohan.xie\}@mbzuai.ac.ae}
}
\begin{document}
\maketitle
\begin{abstract}
In Text-to-SQL tasks, large language models can generate structurally different SQL queries that return correct answers for the same intent. We call this phenomenon \emph{execution-correct structural divergence} (\problem). We introduce \method, a framework that analyzes this behavior through canonical abstract syntax tree representations. We quantify structural diversity and agreement across generations. Experiments with different LLMs on multiple Text-to-SQL datasets, including Spider, document \problem across models and datasets. Furthermore, our experiments demonstrate that generated queries are sensitive to question paraphrases and schema presentation. To address \problem, we adopt a pipeline that first generates structured intermediate representations and then deterministically compiles them into SQL, improving execution accuracy and structural agreement among correct outputs. Structural analysis thus provides an additional diagnostic perspective that complements execution-based evaluation. \mbox{Code is available at \href{https://xanderzhou2022.github.io/AACL2026-SQLSTRUCTEVAL/}{\textcolor{black}{\faGithub\ GitHub}}}.
\end{abstract}

\section{Introduction}
\begin{figure}[t]
\centering
\includegraphics[width=\linewidth]{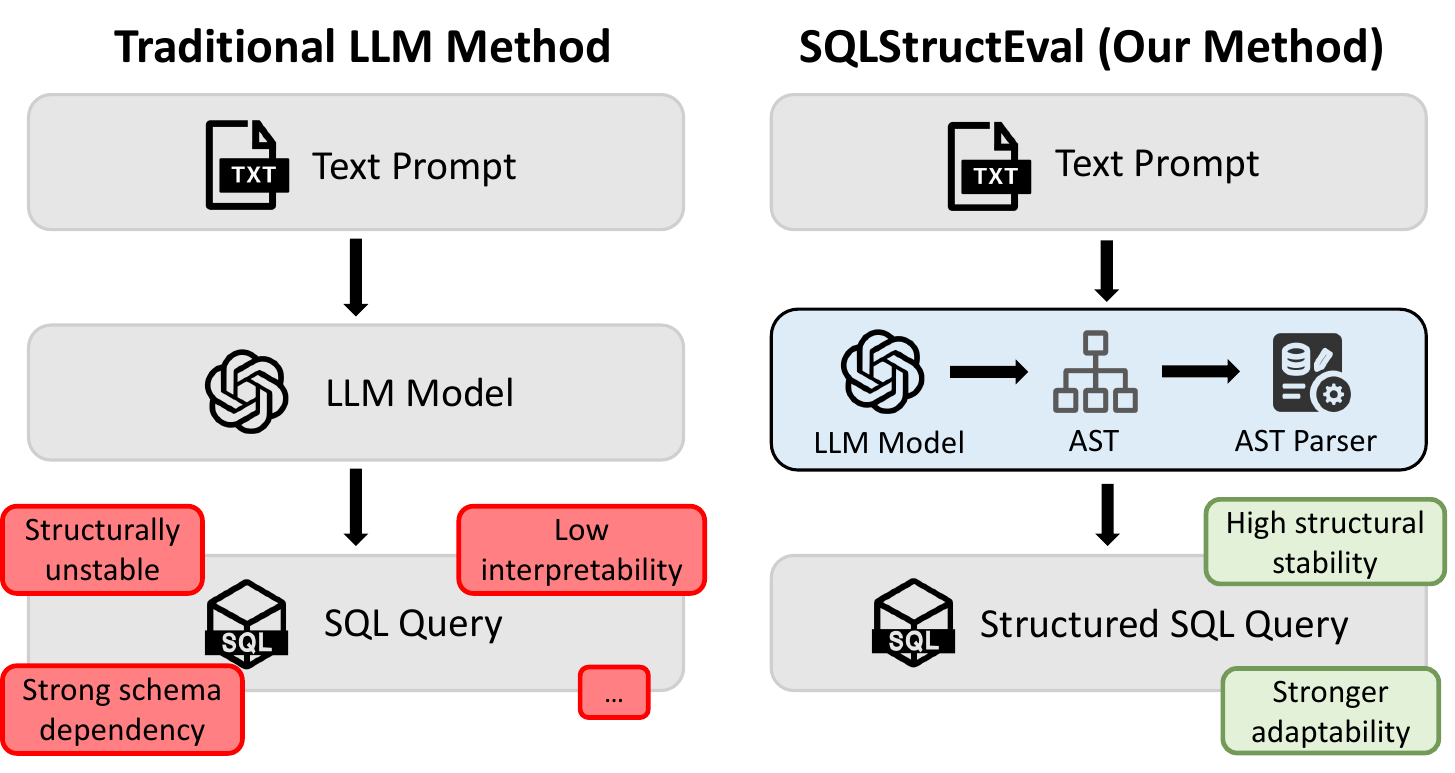}
\caption{Motivation for \method. Repeated Text-to-SQL generation can produce different query constructions even when they yield the same execution result. An AST-based view makes these structural differences explicit, supporting comparison of their diversity, stability, and sensitivity to input variations alongside conventional execution accuracy.}
\label{fig:framework_comparison}
\end{figure}
Large language models (LLMs) support program generation from natural language, including code synthesis and Text-to-SQL translation~\citep{chen2021codex,austin2021program,rajkumar2022evaluating}. Evaluation commonly focuses on functional correctness: unit tests check generated code, and execution accuracy checks whether SQL produces the reference result on a database~\citep{yu2018spider,zhong2020semantic}. These tests are essential for evaluating whether a program answers the intended question. They leave open a different question: how consistently does a model construct that program across repeated generations and equivalent presentations of the input?

Programs that return the same result can use different joins, nested subqueries, aggregation arguments, or redundant operators. Comparing these constructions reveals how a model answers the question. Some alternatives remain equivalent across databases; others produce matching results on the evaluated instance and diverge as the data changes. A model can also change its preferred formulation when the question is reworded or the schema elements are reordered. We study these patterns across repeated generations. We call the occurrence of at least two queries that match the reference result for the same natural language intent but have different canonical ASTs \emph{execution-correct structural divergence} (\problem).

We introduce \method to summarize canonical SQL structures across repeated generations. It measures concentration, diversity, reference alignment, and agreement under input perturbations. Figure~\ref{fig:framework_comparison} illustrates this structural perspective. We analyze execution-correct outputs separately to characterize \problem and include a compile-style generation case study.

Structural reliability describes how consistently a model constructs queries across repeated generations and equivalent inputs. Examining these constructions alongside execution results reveals recurring choices and alternatives for further semantic inspection.

In experiments with seven LLMs on Spider, we observe \problem across model families. We also find that perturbing the natural language expression of a user's intent can change the generated structure. To examine the scope of \problem, we conduct experiments on three subsets of 100 examples each, which also demonstrate its presence. Finally, a compile-style pipeline improves execution accuracy and agreement among correct outputs in our setting.

Our contributions are as follows:
\begin{itemize}
\item We identify the phenomenon of \problem in LLM-generated SQL, showing how correct answers arise from different query constructions.
\item We introduce \method, a framework for measuring structural concentration, diversity, reference alignment, and perturbation sensitivity through canonical AST representations.
\item We document \problem across model families and additional dataset subsets, and evaluate compile-style generation as a full-pipeline mitigation case study.
\end{itemize}

\section{Related Work}
\paragraph{Variation in model generations.}
Self-consistency aggregates multiple sampled reasoning paths~\citep{wang2023selfconsistency}; related work studies the propagation of early errors~\citep{zhang2023how}, and limits of self-correction~\citep{huang2024large}. Instance-level randomization addresses variation in evaluation scores arising from random experimental factors~\citep{li2025instance}.

Program structure has also received direct attention. \citet{song2025structural} analyze code-generation stability through AST-based structural entropy, while \citet{rajput2025dynamic} study algorithmic and runtime variation through opcode distributions. We connect SQL structure with execution correctness and input robustness.

\paragraph{Execution and semantic evaluation.}
Text-to-SQL evaluation distinguishes generalization to new query patterns from robustness to reformulations~\citep{finegan2018evaluation}; cross-database tests reveal further transfer challenges~\citep{suhr2020generalization}. Benchmarks and surveys document execution and reference-based evaluation~\citep{yu2018spider,deng2022recent}. Stronger tests expose errors missed by original tests~\citep{liu2023is}. Distilled SQL test suites probe equivalence across database instances~\citep{zhong2020semantic}, while FLEX studies false execution-based judgments~\citep{kim2024flex}. These approaches assess correctness; \method characterizes \problem alongside broader structural variation.

\paragraph{Robustness benchmarks.}
Spider-Syn tests synonym substitutions~\citep{gan2021syn}; Dr.Spider perturbs questions, databases, and SQL~\citep{chang2023drspider}. ADVETA examines adversarial table perturbations~\citep{pi2022adveta}. BIRD emphasizes database-grounded queries~\citep{li2023bird}, while Spider~2.0 targets enterprise workflows involving heterogeneous dialects and metadata~\citep{lei2025spider2}.

\paragraph{Structured program generation.}
IRNet generates SemQL before constructing SQL~\citep{guo2019complex}; RAT-SQL models relations for schema encoding and linking~\citep{wang2020ratsql}. DIN-SQL decomposes prompting~\citep{Pourreza2023dinsql}; DAIL-SQL studies prompt representations and example selection~\citep{gao2024dailsql}. LEVER verifies language-to-code candidates using execution~\citep{ni2023lever}. Synchromesh and PICARD constrain generation using program structure or incremental parsing~\citep{poesia2022synchromesh,scholak2021picard}. Grammar-constrained decoding further supports structured outputs and logical parsing~\citep{geng2023grammar,raspanti2025grammar}; other methods incorporate context-sensitive or logical constraints~\citep{albinhassan2025learning,ma2025logically}.

PTD-SQL uses query-type-specific demonstrations~\citep{luo2024ptd}; SENSE learns from strong-model examples and weak-model errors~\citep{yang2024sense}. CHASE-SQL combines diverse reasoning paths with pairwise candidate selection~\citep{pourreza2025chase}. These methods connect candidate diversity with answer quality. Our evaluation follows this connection through the SQL constructions themselves: which choices recur, which produce correct answers, and which change when the input is rephrased. This provides a shared view for examining outputs from direct generation and pipelines that first organize query components.

\clearpage
\twocolumn[{
\begin{minipage}{\textwidth}
\centering
\captionsetup{hypcap=false}
\includegraphics[width=\textwidth]{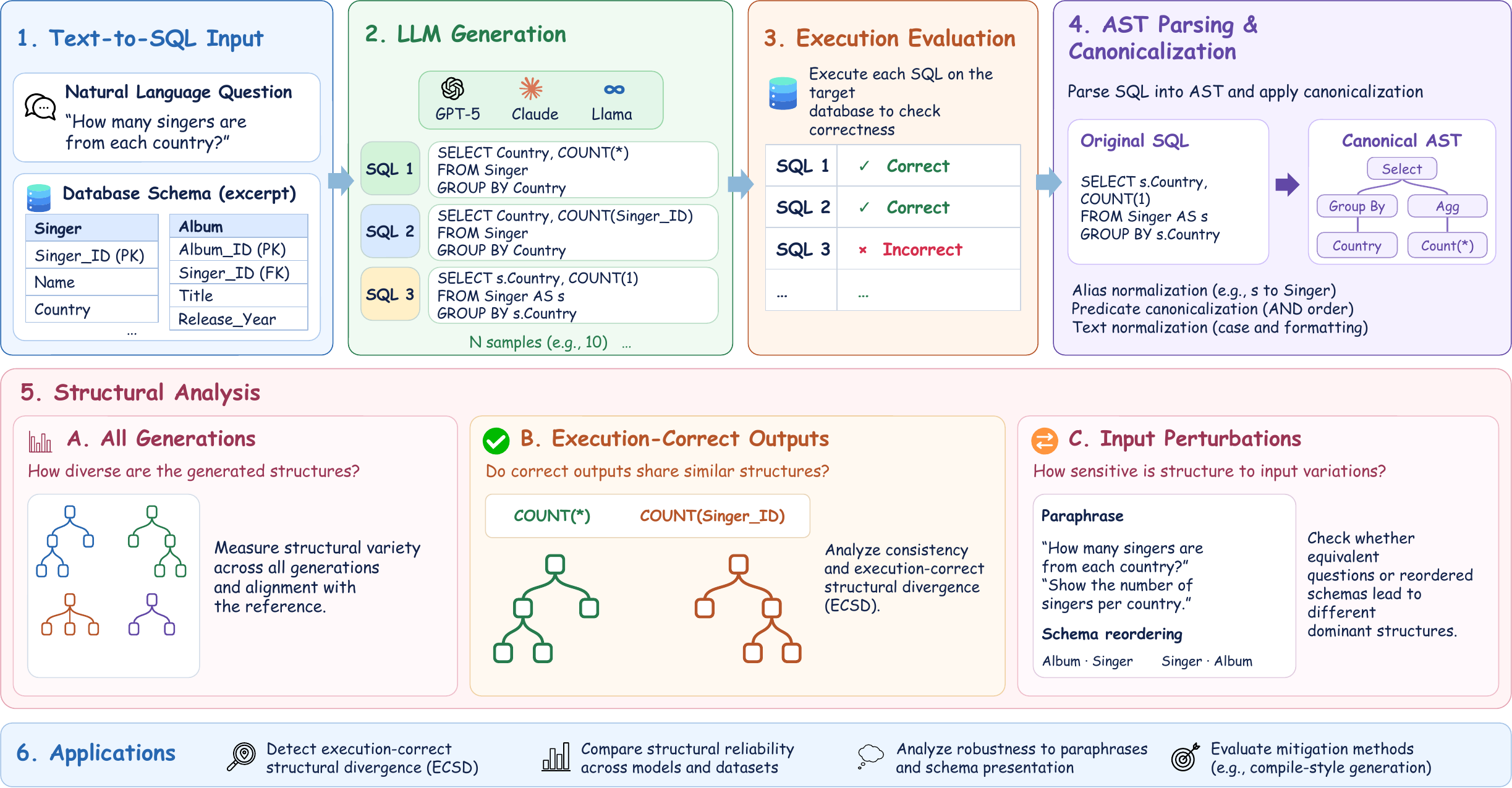}
\captionof{figure}{Overview of the \method evaluation pipeline. Repeated SQL generations for a question and schema undergo execution checks and structural canonicalization. We compare structures across all parsed outputs, within the execution-correct subset, and across input variants. These views connect answer accuracy with structural diversity, agreement, and perturbation sensitivity.}
\label{fig:evaluation_pipeline}
\end{minipage}
\vspace{1em}
}]

\section{\method}
\label{sqlstructeval}
Figure~\ref{fig:evaluation_pipeline} brings together the two assessment paths and the structural comparisons developed in this section.
\subsection{Problem Setup}
\label{problem-setup}
Given a natural language question $x$ and database schema $S$, a Text-to-SQL model generates a query intended to answer $x$. We sample $N$ candidate queries $\{y_i\}_{i=1}^{N}$ for the same input. Repeated generation lets us follow both the answers a model produces and the query constructions it chooses. For example, when asked to count singers by country, a model may generate a grouped query using \texttt{COUNT(*)} in one attempt and \texttt{COUNT(Singer\_ID)} in another. Both can return the reference result on the evaluated database, while expressing the counting operation differently.

We assess each candidate from two perspectives. The first asks whether the query answers the question correctly on the given database. We execute the candidate and compare its result with the annotated reference result. Repeating this check across the $N$ candidates tells us how frequently the model answers correctly and identifies the correct outputs for further analysis.

The second asks how the model constructs its answers. We convert each query into a common structural representation, described in Section~\ref{structural-representation}, and group candidates that share this representation. We first examine all parsed outputs to see the range of constructions the model produces. We then examine the execution-correct subset to see whether correct answers come from a recurring construction or several alternatives. This second view is where we identify \problem. Section~\ref{Structural-Metrics} develops the measures for these comparisons and extends them to question paraphrases and schema reordering. Together, the two assessments connect answer accuracy with the variety and stability of the queries that produce those answers.

\subsection{Structural Representation}
\label{structural-representation}
Comparing query constructions begins with a common way to represent them. Two SQL strings can differ in spacing or alias names while organizing the same operations in the same way. We standardize these details so that repeated constructions can be grouped together. At the same time, we retain choices such as introducing a subquery or changing an aggregation argument, which help explain how the model builds an answer.

An abstract syntax tree (AST) represents this organization through operators such as \texttt{SELECT}, \texttt{JOIN}, \texttt{WHERE}, and \texttt{GROUP BY}, together with their arguments and relationships. For singer counts, the tree records the grouping column and the \texttt{COUNT} argument, preserving the two counting expressions.

We parse SQL using \texttt{sqlglot} in the SQLite dialect. We then rename table aliases by first appearance, update column references, and flatten and sort conjunctions. For example, reordering the conditions in an \texttt{AND} expression is standardized by this step. Finally, we render the AST into SQL and normalize whitespace, case, trailing semicolons, and simple column aliases. Appendix~\ref{app:implementation} gives the implementation details.

The resulting canonical SQL string is our comparison representation: two queries share a structure when their normalized strings match. We can therefore group repeated constructions and count how often each appears. For every generation paradigm, this procedure applies to the final SQL, including SQL compiled from JSON.

\subsection{Structural Evaluation Measures}
\label{Structural-Metrics}
We organize the measures around a sequence of questions: which constructions recur, how they relate to the reference query, which ones produce correct answers, and how they change when the input is rephrased or reordered. Throughout, the basic unit of analysis is the collection of queries generated for one question.

For a question $x$, parsing maps generation $i$ to a canonical structure $a_i\in\mathcal{A}$ or a failure $a_i=\bot$. The valid index set $\mathcal{I}_x=\{i:a_i\ne\bot\}$ contains the outputs available for structural comparison, and $M_x=|\mathcal{I}_x|$ is their number. Among them, let $s_1,\ldots,s_K$ be the distinct structures, with counts $n_k$ and relative frequencies $p_k=n_k/M_x$. These frequencies describe how the sampled outputs are distributed across query constructions.

\paragraph{Concentration and diversity.}
We begin by asking whether the model repeatedly returns to one construction. The majority-structure ratio (MR) is the share of parsed outputs belonging to the most frequent structure:
\begin{equation}
\mathrm{MR}(x)=\max_{1\leq k\leq K}p_k.
\end{equation}
A high MR indicates that one construction dominates the sample. To describe the alternatives around it, we also count the distinct structures (D) and calculate structural entropy (H):
\begin{equation}
\mathrm{D}(x)=K,\qquad
\mathrm{H}(x)=-\sum_{k=1}^{K}p_k\log_2 p_k.
\end{equation}
D counts the constructions that appear, while H describes how evenly outputs are spread across them. Together with MR, they distinguish an occasional alternative from several equally frequent constructions.

We report normalized entropy (NE) as $\mathrm{NE}(x)=\mathrm{H}(x)/\log_2 K$ when $K>1$, and zero when $K\leq1$. Dividing by $\log_2 K$ expresses entropy relative to the maximum for the observed number of structures; NE reaches one when those structures occur equally often. For execution-correct subsets, we report unnormalized entropy ($H_c$).

\paragraph{Annotated-reference alignment.}
Having described the model's recurring constructions, we next compare them with the annotated query. Reference alignment (RA) is the fraction of parsed outputs that match its canonical representation $a^*$:
\begin{equation}
\mathrm{RA}(x)=\frac{1}{M_x}\sum_{i\in\mathcal{I}_x}\mathbf{1}[a_i=a^*].
\end{equation}
RA tells us how often the model follows the reference construction. In the singer-count example, consistently correct queries using a different counting argument would yield high execution accuracy and low RA, revealing a successful alternative that the model repeatedly chooses.

\paragraph{Execution-correct structural agreement.}
We now focus on the outputs that answer the question correctly. This connects the structural distribution directly to \problem: how much variety remains among correct answers? Let $E_x$ contain all execution-correct generation indices and let $C_x=E_x\cap\mathcal{I}_x$ contain those with a parsed structure. For a nonempty selected subset $B_x\subseteq\mathcal{I}_x$, define
\begin{equation}
\mathrm{CA}(B_x)=\frac{\max_{s\in\mathcal{A}}|\{i\in B_x:a_i=s\}|}{|B_x|}.
\end{equation}
Correct-output agreement (CA) applies this measure to $C_x$: it is the share of parsed correct outputs using the most frequent structure. The correct-output distinct count ($D_c$) counts structures in the same subset, and $H_c$ summarizes their frequency distribution. We observe \problem for a question when $D_c>1$.

Comparing correct outputs with the full sample shows whether structural variety occurs among correct answers, incorrect answers, or both. Empty correct subsets receive zero for CA, $D_c$, and $H_c$ and remain in the dataset average.

\paragraph{Diagnostic mismatch indicators.}
To locate these patterns across a dataset, we report two question-level diagnostic rates. Let $e(x)$ be the fraction of the $N$ generations whose execution result matches the reference. The low-agreement rate (LAR) is the fraction of questions with $|E_x|\geq2$ and $\widetilde{\mathrm{CA}}(C_x)<0.7$, where $\widetilde{\mathrm{CA}}$ equals $\mathrm{CA}$ for nonempty subsets and zero otherwise. It highlights questions with dispersed correct structures and also includes cases where all correct outputs fail structural parsing. The high-accuracy diversity rate (HDR) is the fraction with $e(x)\geq0.8$ and $D_c>1$. It locates \problem among questions the model usually answers correctly.

We choose 0.7 and 0.8 as diagnostic thresholds. With ten generations, HDR requires at least eight correct outputs spanning multiple parsed structures. For nonempty correct subsets, LAR selects cases where the most frequent structure accounts for less than 70\%. Together with CA and $D_c$, these rates show how widely correct-answer diversity and low agreement occur.

\paragraph{Robustness under perturbations.}
So far, the input has stayed fixed. We next ask whether the model retains its preferred construction when the question is paraphrased or the schema reordered. We generate queries for each input presentation and compare their most frequent structures.

For an original input $x^{(0)}$ and $T$ variants $x^{(1)},\ldots,x^{(T)}$, let $s_t^*$ denote the majority structure for input $t$, with $t=0$ denoting the original. Cross-variant agreement (VA) is the fraction of input pairs that share a majority structure:
\begin{equation}
\mathrm{VA}(x)=\frac{2}{T(T+1)}\sum_{0\leq u<v\leq T}\mathbf{1}[s_u^*=s_v^*].
\end{equation}
Perturbation sensitivity (Sens) takes the original input as the reference and measures the fraction of variants whose majority changes:
\begin{equation}
\mathrm{Sens}(x)=\frac{1}{T}\sum_{t=1}^{T}\mathbf{1}[s_t^*\ne s_0^*].
\end{equation}
The sensitive fraction (SF) is the proportion of questions with at least one changed majority. The pooled distinct count ($D_p$) counts unique structures across all variants. Together, these measures describe how often the preferred construction changes, how many questions are affected, and the total variety encountered.

\paragraph{Execution, coverage, and reporting.}
We read structural measures alongside execution and parsing results. Execution accuracy (EA) is $e(x)=|E_x|/N$, the share of all generations that answer correctly. Conditional execution accuracy (CEA) divides $|E_x|$ by the number of outputs that execute successfully, using zero when all executions fail. It shows how often a query returns the right answer once it can run. The SQL parse rate (PR) is $M_x/N$, recording how much of the generated sample is available for structural comparison.

We calculate each statistic per question and then average across the dataset, giving every question equal weight. Empty structural subsets contribute zero to counts, concentration, and entropy; reference alignment is also zero when the generated structural subset is empty. Thus, the reported CA and MR averages reflect both structural coverage and concentration. Appendix~\ref{app:measures} summarizes these conventions. In tables, $\uparrow$ marks higher accuracy, coverage, or structural agreement, and $\downarrow$ marks lower perturbation sensitivity. The remaining measures describe the distribution of structures and are shown without directional arrows.

\subsection{Compile-Style Generation}
The measures above describe how query constructions vary. We also examine a procedure that makes these constructions explicit before producing SQL. The model fills a JSON representation with selection, tables, joins, predicates, grouping, ordering, and limits; a deterministic compiler converts these components into executable SQL. Appendix~\ref{app:pipeline} illustrates the sequence from question to JSON, internal AST, and SQL.

For the singer-count question, the model specifies the country column, a counting expression, and the grouping field. The compiler assembles these choices into a SQL query. This intermediate step makes the selected operations and their arguments available for inspection before execution. Across repeated generations, we can then follow which component choices recur and which lead to alternative constructions. The resulting SQL connects these choices back to the evaluation: execution identifies successful answers, while canonicalization groups the constructions used to produce them.

We evaluate compiled and directly generated SQL with the same execution checks and canonicalization. Experiment~3 examines how the pipeline affects accuracy and structural variety among correct answers.

\clearpage
\twocolumn[{
\begin{minipage}{\textwidth}
\centering
\captionsetup{hypcap=false}
\includegraphics[width=\textwidth]{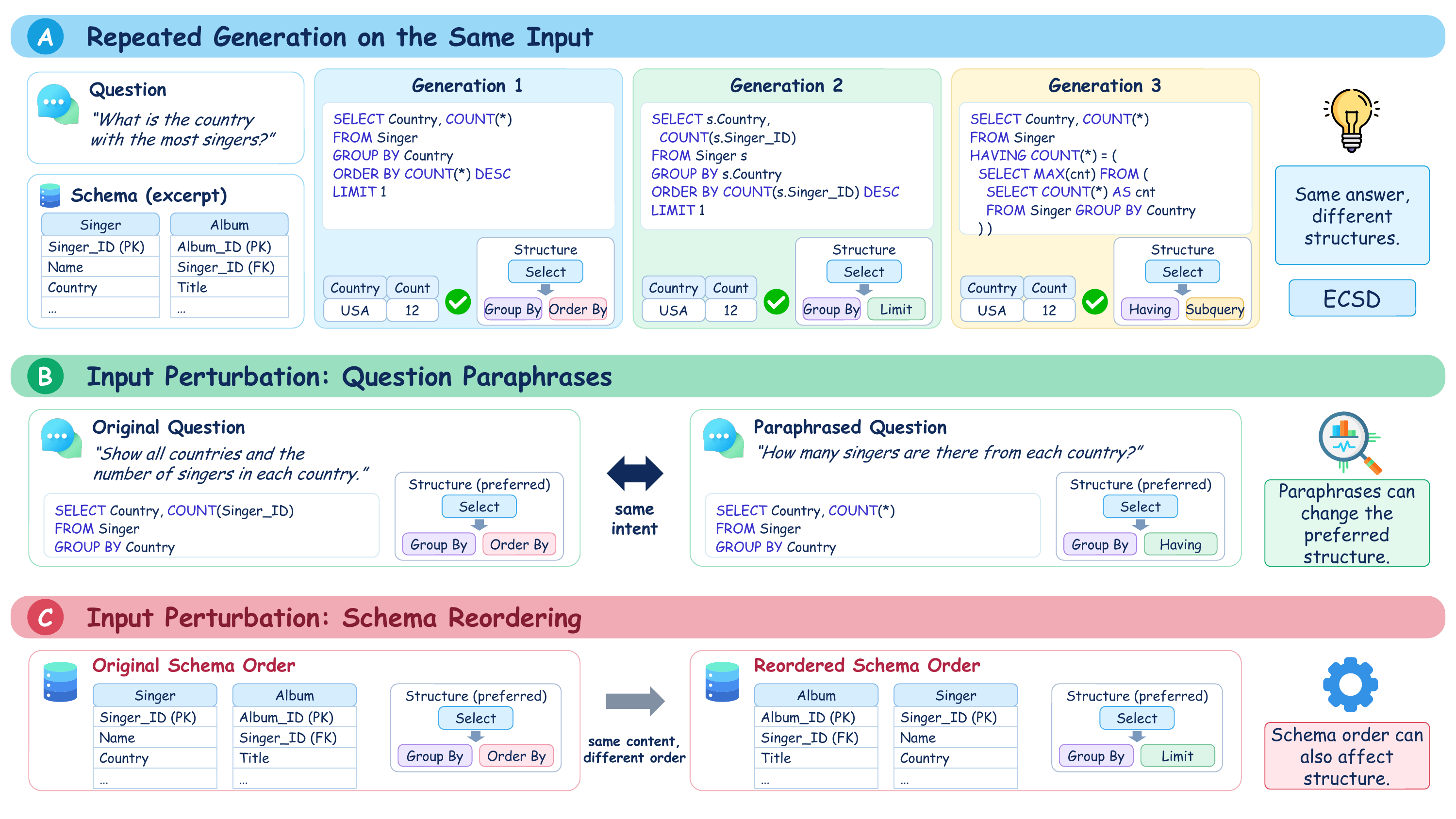}
\captionof{figure}{Structural variation examined in our experiments. (A) Repeated generation for a fixed input can produce different query constructions with matching answers on the evaluated database. (B) Question paraphrases and (C) schema reordering can change the preferred construction. The examples illustrate the comparisons used to study repeated-generation diversity and input sensitivity.}
\label{fig:structure_variation}
\end{minipage}
\vspace{1em}
}]

\section{Experiments and Evaluation}
\label{Experiments and Evaluation}
Figure~\ref{fig:structure_variation} illustrates the structural changes we follow across repeated generations and input variants. We now use the two perspectives from Section~\ref{problem-setup} to follow how models construct answers in practice. The experiments proceed in four steps. We first hold the input fixed and examine the structures that recur across generations. We then check which queries answer correctly and ask how much structural variety remains among them. Next, we evaluate whether compile-style generation changes this combination of accuracy and structural agreement. Finally, we vary the presentation of the input to see how often the model's preferred construction changes. This sequence connects the distributions defined in Section~\ref{Structural-Metrics} to concrete patterns in generated SQL.

\subsection{Experimental Setup}
\paragraph{Datasets.}
Our main experiments use the 1,034-question Spider development set~\citep{yu2018spider}, whose annotated queries and executable databases support joint execution and structural evaluation. Experiment~2 extends the analysis to 100-example subsets of BIRD Mini-Dev, Spider-Syn, and Dr.Spider~\citep{li2023bird,gan2021syn,chang2023drspider}.

\paragraph{Models and sampling.}
We evaluate \modelname{gpt5mini}, \modelname{gpt41mini}, \modelname{claudesonnet}, \modelname{claudeopus}, \modelname{deepseek}, \modelname{geminiflash}, and \modelname{geminipro}. Each input includes the question and database schema, including tables, columns, and foreign keys. We sample ten outputs per input, using temperature 1.0 and default nucleus sampling unless the model API restricts these settings. Appendix~\ref{app:prompts} provides the prompting details.

Repeated sampling gives each question a small collection of query choices. We retain the individual outputs so that their execution results can be matched to their canonical structures. This lets us follow the same collection from overall variation to variation among correct answers, connecting the first two experiments.

\paragraph{Execution and aggregation.}
We execute each query on its SQLite database and compare the result with the reference; execution errors and mismatches count as incorrect. Canonicalization groups parsed outputs by structure, both across all outputs and within the correct subset. We average per-question statistics, assigning zero to empty structural subsets as in Section~\ref{Structural-Metrics}. Each input has one batch of ten generations; a diagnostic rate of 0.01 on a 100-question subset corresponds to one question.

\subsection{Experiment 1: Repeated-Generation Variance}
We first hold the question and schema fixed and generate ten queries per question, yielding 10,340 queries per model. Table~\ref{tab:experiment1_ast_summary} summarizes whether their canonical structures concentrate on one construction or spread across several.

\begin{table}[t]
\centering
\small
\resizebox{\columnwidth}{!}{
\begin{tabular}{lcccc}
\toprule
\rowcolor{tableHeader!65!white}
Model & D & MR$\uparrow$ & NE & RA \\
\midrule
\rowcolor{modelGPT} \modelentry{gpt5mini}  & 1.913 & 0.650 & 0.413 & 0.197 \\
\rowcolor{modelGPT} \modelentry{gpt41mini}  & 1.620 & 0.793 & 0.301 & 0.253 \\
\rowcolor{modelClaude} \modelentry{claudeopus}  & 0.779 & 0.687 & 0.049 & 0.391 \\
\rowcolor{modelClaude} \modelentry{claudesonnet}  & 1.241 & 0.775 & 0.195 & 0.356 \\
\rowcolor{modelDeepSeek} \modelentry{deepseek}  & 1.023 & 0.735 & 0.136 & 0.321 \\
\rowcolor{modelGemini} \modelentry{geminipro}  & 0.845 & 0.595 & 0.133 & 0.369 \\
\rowcolor{modelGemini} \modelentry{geminiflash}  & 0.623 & 0.556 & 0.044 & 0.332 \\
\bottomrule
\end{tabular}
}
\caption{Repeated SQL generation on Spider yields different structural distributions. \modelname{gpt5mini} has the largest D and NE, while \modelname{gpt41mini} has the highest MR.}
\label{tab:experiment1_ast_summary}
\end{table}

\modelname{gpt5mini} produces the most distinct structures, 1.913 per question, and the highest normalized entropy, 0.413. \modelname{gpt41mini} has a higher majority ratio, 0.793 versus 0.650, with lower diversity and entropy. These patterns distinguish a broader set of alternatives from a stronger tendency to repeat one construction.

For \modelname{gpt41mini}, MR 0.793 and RA 0.253 show that a recurring construction can differ from the annotation. This leads to the next question: which constructions answer correctly? Empty parsed subsets contribute zero, explaining the below-one average distinct counts for several models.

\subsection{Experiment 2: Validating \problem}
We execute the same queries from Experiment~1 and examine the correct subset (Table~\ref{tab:experiment2_exec_structure}). This follows the full structural distribution into the outputs that successfully answer the question.

\begin{table*}[t]
\centering
\small
\begin{tabularx}{\textwidth}{l*{6}{>{\centering\arraybackslash}X}}
\toprule
\rowcolor{tableHeader!65!white}
Model & EA$\uparrow$ & CEA$\uparrow$ & $D_c$ & CA$\uparrow$ & HDR & LAR \\
\midrule
\rowcolor{modelGPT} \modelentry{gpt5mini} & 0.741 & 0.741 & 1.378 & 0.552 & 0.351 & 0.297 \\
\rowcolor{modelGPT} \modelentry{gpt41mini} & 0.754 & 0.757 & 1.129 & 0.656 & 0.250 & 0.163 \\
\rowcolor{modelClaude} \modelentry{claudesonnet} & 0.757 & 0.804 & 0.714 & 0.624 & 0.054 & 0.201 \\
\rowcolor{modelClaude} \modelentry{claudeopus} & 0.816 & 0.816 & 0.650 & 0.600 & 0.038 & 0.224 \\
\rowcolor{modelDeepSeek} \modelentry{deepseek} & 0.771 & 0.771 & 0.774 & 0.612 & 0.101 & 0.191 \\
\rowcolor{modelGemini} \modelentry{geminiflash} & 0.638 & 0.743 & 0.509 & 0.484 & 0.016 & 0.191 \\
\rowcolor{modelGemini} \modelentry{geminipro} & \textbf{0.842} & 0.843 & 0.721 & 0.537 & 0.139 & 0.337 \\
\bottomrule
\end{tabularx}
\caption{Execution accuracy and structural agreement differ across models on Spider. \modelname{geminipro} has the highest EA, whereas \modelname{gpt41mini} has the highest CA; positive HDR values reveal \problem at high accuracy.}
\label{tab:experiment2_exec_structure}
\end{table*}

\modelname{geminipro} has the highest EA, 0.842, while \modelname{gpt41mini} has the highest dataset-average CA, 0.656. Thus, answer frequency and correct-output concentration show different patterns. CEA adds execution context: \modelname{claudesonnet} scores 0.757 over all outputs and 0.804 among queries that execute successfully.

For \modelname{gpt5mini}, the average distinct count remains 1.378 among correct outputs, compared with 1.913 overall. Its HDR of 0.351 means that 35.1\% of questions have at least eight correct outputs spanning multiple parsed structures. Structural variety therefore persists on many questions the model repeatedly answers correctly.

All seven models have positive HDR, documenting \problem across model families. LAR ranges from 0.163 to 0.337, locating dispersed correct structures or missing parsed correct outputs. The rates jointly describe correct-answer diversity.

\paragraph{Validating \problem beyond Spider.}
We repeat the analysis on three 100-example subsets using \modelname{gpt5mini}, \modelname{claudesonnet}, and \modelname{deepseek}, with ten queries per question (Table~\ref{tab:scope}). This checks the same pattern under additional question and database settings.

\begin{table*}[t]
\centering
\small
\begin{tabularx}{\textwidth}{ll*{4}{>{\centering\arraybackslash}X}}
\toprule
\rowcolor{tableHeader!65!white}
Subset & Model & EA$\uparrow$ & CA$\uparrow$ & LAR & HDR\\
\midrule
\rowcolor{modelGPT} BIRD Mini-Dev & \modelentry{gpt5mini} & 0.145 & 0.091 & 0.090 & 0.110\\
\rowcolor{modelClaude}  & \modelentry{claudesonnet} & 0.214 & 0.210 & 0.060 & 0.040\\
\rowcolor{modelDeepSeek}  & \modelentry{deepseek} & 0.246 & 0.212 & 0.170 & 0.090\\
\midrule
\rowcolor{modelGPT} Spider-Syn & \modelentry{gpt5mini} & 0.229 & 0.186 & 0.070 & 0.150\\
\rowcolor{modelClaude}  & \modelentry{claudesonnet} & 0.332 & 0.410 & 0.010 & 0.000\\
\rowcolor{modelDeepSeek}  & \modelentry{deepseek} & 0.355 & 0.300 & 0.220 & 0.100\\
\midrule
\rowcolor{modelGPT} Dr.Spider & \modelentry{gpt5mini} & 0.122 & 0.121 & 0.030 & 0.070\\
\rowcolor{modelClaude}  & \modelentry{claudesonnet} & 0.244 & 0.276 & 0.030 & 0.010\\
\rowcolor{modelDeepSeek}  & \modelentry{deepseek} & 0.266 & 0.287 & 0.110 & 0.060\\
\bottomrule
\end{tabularx}
\caption{\problem is observed on the three additional subsets. HDR is positive for all evaluated settings except \modelname{claudesonnet} on Spider-Syn.}
\label{tab:scope}
\end{table*}

HDR reveals \problem in all three subsets. For \modelname{gpt5mini}, it reaches 0.150 on Spider-Syn, 0.110 on BIRD Mini-Dev, and 0.070 on Dr.Spider. On Spider-Syn, 15 questions thus have frequent correct answers with multiple constructions, alongside a dataset-average EA of 0.229. Appendix~\ref{app:scope} adds coverage and diversity statistics. We next examine how changing the generation process affects these patterns.

\subsection{Experiment 3: A Full-Pipeline Case Study}
The compile-style procedure makes query components explicit before deterministic assembly. We compare it with Direct SQL and the decomposed-prompting method DIN-SQL~\citep{Pourreza2023dinsql} on Spider using \modelname{gpt5mini}, ten outputs per question, and the same decoding configuration. The comparison asks whether this process improves both accuracy and correct-output agreement.

Alongside the shared execution and structural checks, we track valid JSON (JV), successful compilation (CR), and successful execution (ES). These rates and PR divide by all generation attempts before averaging across questions, tracing the path from representation to runnable query.

\begin{table}[t]
\centering
\small
\resizebox{\columnwidth}{!}{
\begin{tabular}{l>{\columncolor{tableLilac!65!white}}c>{\columncolor{tableBlue!27!white}}c>{\columncolor{tableCyan!40!white}}c}
\toprule
\rowcolor{tableHeader!65!white}
Measure & Direct SQL & DIN-SQL & Compile-style \\
\midrule
EA$\uparrow$ & 0.742 & 0.736 & \textbf{0.785} \\
CA$\uparrow$ & 0.552 & 0.579 & \textbf{0.632} \\
D & 1.908 & 1.553 & 2.527 \\
\midrule
JV$\uparrow$ & -- & -- & 0.998 \\
CR$\uparrow$ & -- & -- & 0.971 \\
PR$\uparrow$ & -- & -- & 0.902 \\
ES$\uparrow$ & -- & -- & 0.959 \\
\bottomrule
\end{tabular}
}
\caption{On Spider with \modelname{gpt5mini}, compile-style generation improves EA and CA over both baselines while producing more distinct structures overall.}
\label{tab:compile_comparison}
\end{table}

Compile-style generation improves both outcomes (Table~\ref{tab:compile_comparison}): EA reaches 0.785 versus 0.742 for Direct SQL and 0.736 for DIN-SQL; CA reaches 0.632 versus 0.552 and 0.579. The gains over Direct SQL are 4.3 percentage points in accuracy and 0.080 in average CA, which reflects both correct structural coverage and concentration.

The full sample adds another part of the story. Compile-style generation produces 2.527 distinct structures on average, compared with 1.908 for Direct SQL, while also achieving the higher CA above. The output collection becomes more varied overall as the correct-output agreement measure improves. The two views in Section~\ref{Structural-Metrics} connect overall variety with the constructions that produce correct answers.

The intermediate checks help trace this process. Valid JSON is produced in 99.8\% of attempts, 97.1\% compile, and 95.9\% execute successfully. Structural parsing succeeds for 90.2\%; parsing and database execution are separate checks, so each describes a different part of the output's usability. The examples in Appendices~\ref{appendix:error} and~\ref{app:examples} make the remaining query choices visible, including successful constructions and omitted joins or predicates. These examples connect the aggregate improvements to concrete components that can be inspected in the generated representation.

\subsection{Experiment 4: Input Perturbations}
The preceding experiments follow repeated generation for a fixed input and examine a change to the generation pipeline. We now return to the input side of the framework: when the intended question stays the same but its presentation changes, does the model continue to favor the same SQL construction? This connects the observed sampling distributions to the perturbation measures in Section~\ref{Structural-Metrics}.

We use 200 Spider development questions with two kinds of variants. Question paraphrases change the wording; schema variants reorder tables or columns while preserving the database entities and foreign keys. Candidate paraphrases are generated with an LLM prompt and manually checked to preserve requested attributes, aggregation, filters, comparisons, ordering, limits, and schema meaning. Appendix~\ref{app:paraphrases} gives the prompt and checking criteria.

Repeated generation estimates each input's majority structure. VA compares these majorities, Sens tracks departures from the original, and SF records affected questions. The pooled count $D_p$ describes the range of constructions across variants.

\paragraph{Question paraphrases.}
Rewording a question often changes the sampled majority structure (Table~\ref{tab:exp4_paraphrase}). For \modelname{gpt5mini}, SF is 0.900: 180 of the 200 questions have a changed majority under at least one paraphrase. Its VA of 0.328 shows that agreement between presentations is also relatively low. \modelname{geminipro} retains its preferred construction more often, with VA 0.894 and SF 0.195, corresponding to 39 questions with a change. The same evaluation thus exposes both widespread changes and more persistent structural preferences across models.

\begin{table}[t]
\centering
\small
\resizebox{\columnwidth}{!}{
\begin{tabular}{lcccc}
\toprule
\rowcolor{tableHeader!65!white}
Model & VA$\uparrow$ & $D_p$ & Sens$\downarrow$ & SF$\downarrow$ \\
\midrule
\rowcolor{modelGPT} \modelentry{gpt5mini} & 0.328 & 13.220 & 0.622 & 0.900 \\
\rowcolor{modelGPT} \modelentry{gpt41mini} & 0.493 & 6.640 & 0.462 & 0.770 \\
\rowcolor{modelDeepSeek} \modelentry{deepseek} & 0.655 & 3.330 & 0.280 & 0.515 \\
\rowcolor{modelClaude} \modelentry{claudesonnet} & 0.755 & 2.190 & 0.179 & 0.365 \\
\rowcolor{modelClaude} \modelentry{claudeopus} & 0.755 & 2.190 & 0.179 & 0.365 \\
\rowcolor{modelGemini} \modelentry{geminiflash} & 0.787 & 2.870 & 0.158 & 0.395 \\
\rowcolor{modelGemini} \modelentry{geminipro} & \textbf{0.894} & 1.530 & \textbf{0.079} & \textbf{0.195} \\
\bottomrule
\end{tabular}
}
\caption{Question paraphrases change dominant SQL structures across models. \modelname{gpt5mini} has the highest SF, while \modelname{geminipro} has the highest VA.}
\label{tab:exp4_paraphrase}
\end{table}

The pooled samples show the variety accompanying these shifts. Across paraphrases, an average question elicits 13.220 distinct structures from \modelname{gpt5mini} and 1.530 from \modelname{geminipro}. Their Sens values, 0.622 and 0.079, show how often the majority changes from the original input. Together, $D_p$ and Sens describe the range of constructions and the persistence of the model's preferred choice.

\paragraph{Schema presentation.}
We then keep the question wording fixed and change the order in which the schema is shown. This tests a second source of presentation changes within the same task. Table~\ref{tab:exp4_schema} shows higher VA than the paraphrase experiment for every model. For \modelname{gpt5mini}, VA rises from 0.328 to 0.490 and SF falls from 0.900 to 0.640. For \modelname{geminipro}, VA rises from 0.894 to 0.957 and SF falls from 0.195 to 0.065. Schema reordering thus preserves the preferred construction more often than question paraphrasing.

\begin{table}[t]
\centering
\small
\resizebox{\columnwidth}{!}{
\begin{tabular}{lcccc}
\toprule
\rowcolor{tableHeader!65!white}
Model & VA$\uparrow$ & $D_p$ & Sens$\downarrow$ & SF$\downarrow$ \\
\midrule
\rowcolor{modelGPT} \modelentry{gpt5mini} & 0.490 & 7.330 & 0.500 & 0.640 \\
\rowcolor{modelGPT} \modelentry{gpt41mini} & 0.727 & 3.600 & 0.263 & 0.360 \\
\rowcolor{modelDeepSeek} \modelentry{deepseek} & 0.808 & 1.870 & 0.170 & 0.260 \\
\rowcolor{modelGemini} \modelentry{geminiflash} & 0.895 & 1.790 & 0.098 & 0.145 \\
\rowcolor{modelClaude} \modelentry{claudesonnet} & 0.955 & 1.130 & 0.043 & 0.065 \\
\rowcolor{modelClaude} \modelentry{claudeopus} & 0.955 & 1.130 & 0.043 & 0.065 \\
\rowcolor{modelGemini} \modelentry{geminipro} & \textbf{0.957} & 1.060 & \textbf{0.043} & \textbf{0.065} \\
\bottomrule
\end{tabular}
}
\caption{Schema reordering yields higher VA than question paraphrasing for every evaluated model. Each model exhibits majority-structure changes on some questions.}
\label{tab:exp4_schema}
\end{table}

Schema reordering changes the majority for 128 questions with \modelname{gpt5mini} and 13 with \modelname{geminipro}. Both perturbations thus expose input-sensitive query choices. Together with the earlier experiments, these alternatives connect repeated-generation diversity and correct answers with the semantic inspection discussed next.

\section{Discussion}
The effect of \problem depends on the data: tied minima can separate \texttt{LIMIT 1} from a \texttt{MIN} subquery, while counting rows and non-null identifiers gives the same result. These differences guide semantic tests involving ties, nulls, and duplicates~\citep{guagliardo2017semantics,chu2018equivalence,zhong2020semantic}.

Reading accuracy, correct-output diversity, and perturbation sensitivity together helps identify query alternatives for inspection~\citep{ribeiro2020checklist}. The compile-style results illustrate this joint view: accuracy and correct-output agreement improve alongside greater overall diversity.

Coverage and sampling shape these comparisons. Empty correct subsets contribute zero to CA, so PR and $D_c$ provide useful context. Matched sampling budgets and independent repeats help clarify recurring structural patterns.

\section{Conclusion and Future Work}
We introduced \method to connect execution accuracy with the structure of generated SQL. Experiments revealed \problem across models and datasets, structural changes under input perturbations, and improved accuracy and correct-output agreement with compile-style generation. These findings show how structural analysis makes the routes to a correct answer visible. Following these routes across samples and input variants helps identify recurring query choices and concrete alternatives for closer inspection. Future work will test database changes and separate generation and compilation effects.

\clearpage
\section*{Limitations}
Our main evaluation uses Spider, with additional 100-question subsets from three benchmarks and ten generations per input. Larger datasets, repeated sampling, and threshold comparisons would help assess how consistently the observed patterns extend across settings. Structural statistics depend on parsing coverage, while execution correctness is measured on the supplied database instances. Further semantic tests and component ablations would clarify the practical effects of query alternatives and the contributions of the compile-style pipeline.

\section*{Ethical Considerations}
We evaluate generated queries offline using publicly available Text-to-SQL benchmarks. In practical applications, incorrect queries can lead to misleading decisions. Structural diagnostics can guide inspection, alongside execution tests and semantic checks suited to the application's data and requirements.

\paragraph{Data and artifact use.}
We use Spider, BIRD Mini-Dev, Spider-Syn, and Dr.Spider for research evaluation under their respective access and licensing terms. Derived question variants and generated queries remain associated with those source resources. The code link provides the evaluation implementation; use and redistribution of datasets and model outputs must respect the original resource and provider terms. Hosted model updates can affect reproducibility even when the prompt remains fixed.

\paragraph{AI assistance.}
For manuscript preparation, we used AI solely for language polishing.

\bibliography{refs}

@inproceedings{Pourreza2023dinsql,
  author = {Pourreza, Mohammadreza and Rafiei, Davood},
  title = {{DIN-SQL}: Decomposed In-Context Learning of Text-to-{SQL} with Self-Correction},
  booktitle = {Advances in Neural Information Processing Systems},
  series = {NeurIPS~'23},
  volume = {36},
  pages = {36339--36348},
  address = {New Orleans, LA, USA},
  publisher = {Curran Associates, Inc.},
  year = {2023},
  doi = {10.52202/075280-1577},
  url = {https://proceedings.neurips.cc/paper_files/paper/2023/file/72223cc66f63ca1aa59edaec1b3670e6-Paper-Conference.pdf},
}

@inproceedings{albinhassan2025learning,
  author = {Albinhassan, Mohammad and Madhyastha, Pranava and Law, Mark and Russo, Alessandra},
  title = {Learning and Enforcing Context-Sensitive Control for {LLM}s},
  booktitle = {Proceedings of the 63rd Annual Meeting of the Association for Computational Linguistics (Volume 4: Student Research Workshop)},
  series = {ACL-SRW~'25},
  pages = {834--842},
  address = {Vienna, Austria},
  publisher = {Association for Computational Linguistics},
  year = {2025},
  doi = {10.18653/v1/2025.acl-srw.59},
  url = {https://aclanthology.org/2025.acl-srw.59/},
}

@article{austin2021program,
  author = {Jacob Austin and Augustus Odena and Maxwell Nye and Maarten Bosma and Henryk Michalewski and David Dohan and Ellen Jiang and Carrie Cai and Michael Terry and Quoc Le and Charles Sutton},
  title = {Program Synthesis with Large Language Models},
  journal = {{arXiv} preprint},
  volume = {arXiv:2108.07732},
  year = {2021},
  url = {https://arxiv.org/abs/2108.07732},
}

@inproceedings{chang2023drspider,
  author = {Shuaichen Chang and Jun Wang and Mingwen Dong and Lin Pan and Henghui Zhu and Alexander Hanbo Li and Wuwei Lan and Sheng Zhang and Jiarong Jiang and Joseph Lilien and Steve Ash and William Wang and Zhiguo Wang and Vittorio Castelli and Bing Xiang and Patrick Ng},
  title = {{Dr.Spider}: A Diagnostic Evaluation Benchmark towards Text-to-{SQL} Robustness},
  booktitle = {Proceedings of the Eleventh International Conference on Learning Representations},
  series = {ICLR~'23},
  address = {Kigali, Rwanda},
  publisher = {OpenReview.net},
  year = {2023},
  url = {https://iclr.cc/virtual/2023/poster/11467},
}

@article{chen2021codex,
  author = {Mark Chen and Jerry Tworek and Heewoo Jun and Qiming Yuan and Henrique Ponde de Oliveira Pinto and Jared Kaplan and Harri Edwards and Yuri Burda and Nicholas Joseph and Greg Brockman and Alex Ray and Raul Puri and Gretchen Krueger and Michael Petrov and Heidy Khlaaf and Girish Sastry and Pamela Mishkin and Brooke Chan and Scott Gray and Nick Ryder and Mikhail Pavlov and Alethea Power and Lukasz Kaiser and Mohammad Bavarian and Clemens Winter and Philippe Tillet and Felipe Petroski Such and Dave Cummings and Matthias Plappert and Fotios Chantzis and Elizabeth Barnes and Ariel Herbert{-}Voss and William Hebgen Guss and Alex Nichol and Alex Paino and Nikolas Tezak and Jie Tang and Igor Babuschkin and Suchir Balaji and Shantanu Jain and William Saunders and Christopher Hesse and Andrew N. Carr and Jan Leike and Josh Achiam and Vedant Misra and Evan Morikawa and Alec Radford and Matthew Knight and Miles Brundage and Mira Murati and Katie Mayer and Peter Welinder and Bob McGrew and Dario Amodei and Sam McCandlish and Ilya Sutskever and Wojciech Zaremba},
  title = {Evaluating Large Language Models Trained on Code},
  journal = {{arXiv} preprint},
  volume = {arXiv:2107.03374},
  year = {2021},
  url = {https://arxiv.org/abs/2107.03374},
}

@inproceedings{deng2022recent,
  author = {Deng, Naihao and Chen, Yulong and Zhang, Yue},
  title = {Recent Advances in Text-to-{SQL}: A Survey of What We Have and What We Expect},
  booktitle = {Proceedings of the 29th International Conference on Computational Linguistics},
  series = {COLING~'22},
  pages = {2166--2187},
  address = {Gyeongju, Republic of Korea},
  publisher = {International Committee on Computational Linguistics},
  year = {2022},
  url = {https://aclanthology.org/2022.coling-1.190/},
}

@inproceedings{gan2021syn,
  author = {Gan, Yujian and Chen, Xinyun and Huang, Qiuping and Purver, Matthew and Woodward, John R. and Xie, Jinxia and Huang, Pengsheng},
  title = {Towards Robustness of Text-to-{SQL} Models against Synonym Substitution},
  booktitle = {Proceedings of the 59th Annual Meeting of the Association for Computational Linguistics and the 11th International Joint Conference on Natural Language Processing (Volume 1: Long Papers)},
  series = {ACL-IJCNLP~'21},
  pages = {2505--2515},
  address = {Online},
  publisher = {Association for Computational Linguistics},
  year = {2021},
  doi = {10.18653/v1/2021.acl-long.195},
  url = {https://aclanthology.org/2021.acl-long.195/},
}

@inproceedings{geng2023grammar,
  author = {Geng, Saibo and Josifoski, Martin and Peyrard, Maxime and West, Robert},
  title = {Grammar-Constrained Decoding for Structured {NLP} Tasks without Finetuning},
  booktitle = {Proceedings of the 2023 Conference on Empirical Methods in Natural Language Processing},
  series = {EMNLP~'23},
  pages = {10932--10952},
  address = {Singapore},
  publisher = {Association for Computational Linguistics},
  year = {2023},
  doi = {10.18653/v1/2023.emnlp-main.674},
  url = {https://aclanthology.org/2023.emnlp-main.674/},
}

@inproceedings{guo2019complex,
  author = {Guo, Jiaqi and Zhan, Zecheng and Gao, Yan and Xiao, Yan and Lou, Jian-Guang and Liu, Ting and Zhang, Dongmei},
  title = {Towards Complex Text-to-{SQL} in Cross-Domain Database with Intermediate Representation},
  booktitle = {Proceedings of the 57th Annual Meeting of the Association for Computational Linguistics},
  series = {ACL~'19},
  pages = {4524--4535},
  address = {Florence, Italy},
  publisher = {Association for Computational Linguistics},
  year = {2019},
  doi = {10.18653/v1/P19-1444},
  url = {https://aclanthology.org/P19-1444/},
}

@inproceedings{huang2024large,
  author = {Jie Huang and Xinyun Chen and Swaroop Mishra and Huaixiu Steven Zheng and Adams Wei Yu and Xinying Song and Denny Zhou},
  title = {Large Language Models Cannot Self-Correct Reasoning Yet},
  booktitle = {Proceedings of the Twelfth International Conference on Learning Representations},
  series = {ICLR~'24},
  address = {Vienna, Austria},
  publisher = {OpenReview.net},
  year = {2024},
  url = {https://openreview.net/forum?id=IkmD3fKBPQ},
}

@inproceedings{kim2024flex,
  author = {Kim, Heegyu and Jeon, Taeyang and Choi, Seunghwan and Choi, Seungtaek and Cho, Hyunsouk},
  title = {{FLEX}: Expert-level False-Less {EX}ecution Metric for Reliable Text-to-{SQL} Benchmark},
  booktitle = {Proceedings of the 2025 Conference of the Nations of the Americas Chapter of the Association for Computational Linguistics: Human Language Technologies (Volume 1: Long Papers)},
  series = {NAACL~'25},
  pages = {4448--4475},
  address = {Albuquerque, New Mexico},
  publisher = {Association for Computational Linguistics},
  year = {2025},
  doi = {10.18653/v1/2025.naacl-long.228},
  url = {https://aclanthology.org/2025.naacl-long.228/},
}

@inproceedings{li2023bird,
  author = {Li, Jinyang and Hui, Binyuan and Qu, Ge and Yang, Jiaxi and Li, Binhua and Li, Bowen and Wang, Bailin and Qin, Bowen and Geng, Ruiying and Huo, Nan and Zhou, Xuanhe and Ma, Chenhao and Li, Guoliang and Chang, Kevin C. C. and Huang, Fei and Cheng, Reynold and Li, Yongbin},
  title = {Can {LLM} Already Serve as a Database Interface? A {BIg} Bench for Large-Scale Database Grounded Text-to-{SQL}s},
  booktitle = {Advances in Neural Information Processing Systems},
  series = {NeurIPS~'23},
  volume = {36},
  pages = {42330--42357},
  address = {New Orleans, LA, USA},
  publisher = {Curran Associates, Inc.},
  year = {2023},
  doi = {10.52202/075280-1835},
  url = {https://proceedings.neurips.cc/paper_files/paper/2023/file/83fc8fab1710363050bbd1d4b8cc0021-Paper-Datasets_and_Benchmarks.pdf},
}

@inproceedings{li2025instance,
  author = {Li, Yiyang and Wu, Yonghuang and Luo, Ying and Sun, Liangtai and Qin, Zishu and Qiu, Lin and Cao, Xuezhi and Cai, Xunliang},
  title = {Instance-level Randomization: Toward More Stable {LLM} Evaluations},
  booktitle = {Findings of the Association for Computational Linguistics: EMNLP 2025},
  series = {Findings-EMNLP~'25},
  pages = {3411--3425},
  address = {Suzhou, China},
  publisher = {Association for Computational Linguistics},
  year = {2025},
  doi = {10.18653/v1/2025.findings-emnlp.182},
  url = {https://aclanthology.org/2025.findings-emnlp.182/},
}

@inproceedings{liu2023is,
  author = {Liu, Jiawei and Xia, Chunqiu Steven and Wang, Yuyao and Zhang, Lingming},
  title = {Is Your Code Generated by {ChatGPT} Really Correct? Rigorous Evaluation of Large Language Models for Code Generation},
  booktitle = {Advances in Neural Information Processing Systems},
  series = {NeurIPS~'23},
  volume = {36},
  pages = {21558--21572},
  address = {New Orleans, LA, USA},
  publisher = {Curran Associates, Inc.},
  year = {2023},
  doi = {10.52202/075280-0943},
  url = {https://proceedings.neurips.cc/paper_files/paper/2023/file/43e9d647ccd3e4b7b5baab53f0368686-Paper-Conference.pdf},
}

@inproceedings{ma2025logically,
  author = {Ma, Franklin and Hu, Alan J.},
  title = {Logically Constrained Decoding},
  booktitle = {Proceedings of the 3rd Workshop on Mathematical Natural Language Processing},
  series = {MathNLP~'25},
  pages = {150--167},
  address = {Suzhou, China},
  publisher = {Association for Computational Linguistics},
  year = {2025},
  doi = {10.18653/v1/2025.mathnlp-main.11},
  url = {https://aclanthology.org/2025.mathnlp-main.11/},
}

@inproceedings{ni2023lever,
  author = {Ni, Ansong and Iyer, Srini and Radev, Dragomir and Stoyanov, Ves and Yih, Wen-tau and Wang, Sida I. and Lin, Xi Victoria},
  title = {{LEVER}: Learning to Verify Language-to-Code Generation with Execution},
  booktitle = {Proceedings of the 40th International Conference on Machine Learning},
  series = {Proceedings of Machine Learning Research (ICML~'23)},
  volume = {202},
  pages = {26106--26128},
  address = {Honolulu, HI, USA},
  publisher = {PMLR},
  year = {2023},
  url = {https://proceedings.mlr.press/v202/ni23b.html},
}

@inproceedings{poesia2022synchromesh,
  author = {Gabriel Poesia and Oleksandr Polozov and Vu Le and Ashish Tiwari and Gustavo Soares and Christopher Meek and Sumit Gulwani},
  title = {{Synchromesh}: Reliable Code Generation from Pre-trained Language Models},
  booktitle = {Proceedings of the Tenth International Conference on Learning Representations},
  series = {ICLR~'22},
  address = {Online},
  publisher = {OpenReview.net},
  year = {2022},
  url = {https://openreview.net/forum?id=KmtVD97J43e},
}

@article{rajkumar2022evaluating,
  author = {Nitarshan Rajkumar and Raymond Li and Dzmitry Bahdanau},
  title = {Evaluating the Text-to-{SQL} Capabilities of Large Language Models},
  journal = {{arXiv} preprint},
  volume = {arXiv:2204.00498},
  year = {2022},
  url = {https://arxiv.org/abs/2204.00498},
}

@article{rajput2025dynamic,
  author = {Prateek Rajput and Abdoul Aziz Bonkoungou and Yewei Song and Abdoul Kader Kabore and Iyiola E. Olatunji and Jacques Klein and Tegewende Bissyande},
  title = {Dynamic Stability of {LLM}-Generated Code},
  journal = {{arXiv} preprint},
  volume = {arXiv:2511.07463},
  year = {2025},
  url = {https://arxiv.org/abs/2511.07463},
}

@inproceedings{raspanti2025grammar,
  author = {Raspanti, Federico and Ozcelebi, Tanir and Holenderski, Mike},
  title = {Grammar-Constrained Decoding Makes Large Language Models Better Logical Parsers},
  booktitle = {Proceedings of the 63rd Annual Meeting of the Association for Computational Linguistics (Volume 6: Industry Track)},
  series = {ACL-Industry~'25},
  pages = {485--499},
  address = {Vienna, Austria},
  publisher = {Association for Computational Linguistics},
  year = {2025},
  doi = {10.18653/v1/2025.acl-industry.34},
  url = {https://aclanthology.org/2025.acl-industry.34/},
}

@inproceedings{scholak2021picard,
  author = {Scholak, Torsten and Schucher, Nathan and Bahdanau, Dzmitry},
  title = {{PICARD}: Parsing Incrementally for Constrained Auto-Regressive Decoding from Language Models},
  booktitle = {Proceedings of the 2021 Conference on Empirical Methods in Natural Language Processing},
  series = {EMNLP~'21},
  pages = {9895--9901},
  address = {Online and Punta Cana, Dominican Republic},
  publisher = {Association for Computational Linguistics},
  year = {2021},
  doi = {10.18653/v1/2021.emnlp-main.779},
  url = {https://aclanthology.org/2021.emnlp-main.779/},
}

@inproceedings{song2025structural,
  author = {Song, Yewei and Sun, Tiezhu and Tang, Xunzhu and Rajput, Prateek Kumar and Bissyand{\'{e}}, Tegawend{\'{e}} F. and Klein, Jacques},
  title = {Measuring {LLM} Code Generation Stability via Structural Entropy},
  booktitle = {Proceedings of the 40th IEEE/ACM International Conference on Automated Software Engineering},
  series = {ASE~'25},
  pages = {3922--3926},
  address = {Seoul, Republic of Korea},
  publisher = {IEEE},
  year = {2025},
  doi = {10.1109/ASE63991.2025.00343},
  url = {https://doi.org/10.1109/ASE63991.2025.00343},
}

@inproceedings{wang2023selfconsistency,
  author = {Xuezhi Wang and Jason Wei and Dale Schuurmans and Quoc Le and Ed H. Chi and Sharan Narang and Aakanksha Chowdhery and Denny Zhou},
  title = {Self-Consistency Improves Chain of Thought Reasoning in Language Models},
  booktitle = {Proceedings of the Eleventh International Conference on Learning Representations},
  series = {ICLR~'23},
  address = {Kigali, Rwanda},
  publisher = {OpenReview.net},
  year = {2023},
  url = {https://openreview.net/forum?id=1PL1NIMMrw},
}

@inproceedings{yu2018spider,
  author = {Yu, Tao and Zhang, Rui and Yang, Kai and Yasunaga, Michihiro and Wang, Dongxu and Li, Zifan and Ma, James and Li, Irene and Yao, Qingning and Roman, Shanelle and Zhang, Zilin and Radev, Dragomir R.},
  title = {{S}pider: A Large-Scale Human-Labeled Dataset for Complex and Cross-Domain Semantic Parsing and Text-to-{SQL} Task},
  booktitle = {Proceedings of the 2018 Conference on Empirical Methods in Natural Language Processing},
  series = {EMNLP~'18},
  pages = {3911--3921},
  address = {Brussels, Belgium},
  publisher = {Association for Computational Linguistics},
  year = {2018},
  doi = {10.18653/v1/D18-1425},
  url = {https://aclanthology.org/D18-1425/},
}

@inproceedings{zhang2023how,
  author = {Zhang, Muru and Press, Ofir and Merrill, William and Liu, Alisa and Smith, Noah A.},
  title = {How Language Model Hallucinations Can Snowball},
  booktitle = {Proceedings of the 41st International Conference on Machine Learning},
  series = {Proceedings of Machine Learning Research (ICML~'24)},
  volume = {235},
  pages = {59670--59684},
  address = {Vienna, Austria},
  publisher = {PMLR},
  year = {2024},
  url = {https://proceedings.mlr.press/v235/zhang24ay.html},
}

@inproceedings{zhong2020semantic,
  author = {Zhong, Ruiqi and Yu, Tao and Klein, Dan},
  title = {Semantic Evaluation for Text-to-{SQL} with Distilled Test Suites},
  booktitle = {Proceedings of the 2020 Conference on Empirical Methods in Natural Language Processing},
  series = {EMNLP~'20},
  pages = {396--411},
  address = {Online},
  publisher = {Association for Computational Linguistics},
  year = {2020},
  doi = {10.18653/v1/2020.emnlp-main.29},
  url = {https://aclanthology.org/2020.emnlp-main.29/},
}

@article{chu2018equivalence,
  author = {Chu, Shumo and Murphy, Brendan and Roesch, Jared and Cheung, Alvin and Suciu, Dan},
  title = {Axiomatic Foundations and Algorithms for Deciding Semantic Equivalences of {SQL} Queries},
  journal = {Proceedings of the {VLDB} Endowment},
  volume = {11},
  number = {11},
  pages = {1482--1495},
  year = {2018},
  doi = {10.14778/3236187.3236200},
  url = {https://www.vldb.org/pvldb/vol11/p1482-chu.pdf},
}

@inproceedings{finegan2018evaluation,
  author = {Finegan-Dollak, Catherine and Kummerfeld, Jonathan K. and Zhang, Li and Ramanathan, Karthik and Sadasivam, Sesh and Zhang, Rui and Radev, Dragomir},
  title = {Improving Text-to-{SQL} Evaluation Methodology},
  booktitle = {Proceedings of the 56th Annual Meeting of the Association for Computational Linguistics (Volume 1: Long Papers)},
  series = {ACL~'18},
  address = {Melbourne, Australia},
  publisher = {Association for Computational Linguistics},
  pages = {351--360},
  year = {2018},
  doi = {10.18653/v1/P18-1033},
  url = {https://aclanthology.org/P18-1033/},
}

@article{gao2024dailsql,
  author = {Gao, Dawei and Wang, Haibin and Li, Yaliang and Sun, Xiuyu and Qian, Yichen and Ding, Bolin and Zhou, Jingren},
  title = {Text-to-{SQL} Empowered by Large Language Models: A Benchmark Evaluation},
  journal = {Proceedings of the {VLDB} Endowment},
  volume = {17},
  number = {5},
  pages = {1132--1145},
  year = {2024},
  doi = {10.14778/3641204.3641221},
  url = {https://www.vldb.org/pvldb/vol17/p1132-gao.pdf},
}

@inproceedings{lei2025spider2,
  author = {Lei, Fangyu and Chen, Jixuan and Ye, Yuxiao and Cao, Ruisheng and Shin, Dongchan and Su, Hongjin and Suo, Zhaoqing and Gao, Hongcheng and Hu, Wenjing and Yin, Pengcheng and Zhong, Victor and Xiong, Caiming and Sun, Ruoxi and Liu, Qian and Wang, Sida I. and Yu, Tao},
  title = {{Spider} 2.0: Evaluating Language Models on Real-World Enterprise Text-to-{SQL} Workflows},
  booktitle = {Proceedings of the Thirteenth International Conference on Learning Representations},
  series = {ICLR~'25},
  address = {Singapore},
  publisher = {OpenReview.net},
  year = {2025},
  url = {https://proceedings.iclr.cc/paper_files/paper/2025/hash/46c10f6c8ea5aa6f267bcdabcb123f97-Abstract-Conference.html},
}

@inproceedings{pi2022adveta,
  author = {Pi, Xinyu and Wang, Bing and Gao, Yan and Guo, Jiaqi and Li, Zhoujun and Lou, Jian-Guang},
  title = {Towards Robustness of Text-to-{SQL} Models Against Natural and Realistic Adversarial Table Perturbation},
  booktitle = {Proceedings of the 60th Annual Meeting of the Association for Computational Linguistics (Volume 1: Long Papers)},
  series = {ACL~'22},
  address = {Dublin, Ireland},
  publisher = {Association for Computational Linguistics},
  pages = {2007--2022},
  year = {2022},
  doi = {10.18653/v1/2022.acl-long.142},
  url = {https://aclanthology.org/2022.acl-long.142/},
}

@inproceedings{ribeiro2020checklist,
  author = {Ribeiro, Marco Tulio and Wu, Tongshuang and Guestrin, Carlos and Singh, Sameer},
  title = {Beyond Accuracy: Behavioral Testing of {NLP} Models with {CheckList}},
  booktitle = {Proceedings of the 58th Annual Meeting of the Association for Computational Linguistics},
  series = {ACL~'20},
  address = {Online},
  publisher = {Association for Computational Linguistics},
  pages = {4902--4912},
  year = {2020},
  doi = {10.18653/v1/2020.acl-main.442},
  url = {https://aclanthology.org/2020.acl-main.442/},
}

@inproceedings{suhr2020generalization,
  author = {Suhr, Alane and Chang, Ming-Wei and Shaw, Peter and Lee, Kenton},
  title = {Exploring Unexplored Generalization Challenges for Cross-Database Semantic Parsing},
  booktitle = {Proceedings of the 58th Annual Meeting of the Association for Computational Linguistics},
  series = {ACL~'20},
  address = {Online},
  publisher = {Association for Computational Linguistics},
  pages = {8372--8388},
  year = {2020},
  doi = {10.18653/v1/2020.acl-main.742},
  url = {https://aclanthology.org/2020.acl-main.742/},
}

@inproceedings{wang2020ratsql,
  author = {Wang, Bailin and Shin, Richard and Liu, Xiaodong and Polozov, Oleksandr and Richardson, Matthew},
  title = {{RAT-SQL}: Relation-Aware Schema Encoding and Linking for Text-to-{SQL} Parsers},
  booktitle = {Proceedings of the 58th Annual Meeting of the Association for Computational Linguistics},
  series = {ACL~'20},
  address = {Online},
  publisher = {Association for Computational Linguistics},
  pages = {7567--7578},
  year = {2020},
  doi = {10.18653/v1/2020.acl-main.677},
  url = {https://aclanthology.org/2020.acl-main.677/},
}

@article{guagliardo2017semantics,
  author = {Guagliardo, Paolo and Libkin, Leonid},
  title = {A Formal Semantics of {SQL} Queries, Its Validation, and Applications},
  journal = {Proceedings of the {VLDB} Endowment},
  volume = {11},
  number = {1},
  pages = {27--39},
  year = {2017},
  doi = {10.14778/3151113.3151116},
  url = {https://www.vldb.org/pvldb/vol11/p27-guagliardo.pdf},
}

@inproceedings{luo2024ptd,
  author = {Luo, Ruilin and Wang, Liyuan and Lin, Binghuai and Lin, Zicheng and Yang, Yujiu},
  title = {{PTD-SQL}: Partitioning and Targeted Drilling with {LLM}s in Text-to-{SQL}},
  booktitle = {Proceedings of the 2024 Conference on Empirical Methods in Natural Language Processing},
  series = {EMNLP~'24},
  pages = {3767--3799},
  address = {Miami, Florida, USA},
  publisher = {Association for Computational Linguistics},
  year = {2024},
  doi = {10.18653/v1/2024.emnlp-main.221},
  url = {https://aclanthology.org/2024.emnlp-main.221/},
}

@inproceedings{yang2024sense,
  author = {Yang, Jiaxi and Hui, Binyuan and Yang, Min and Yang, Jian and Lin, Junyang and Zhou, Chang},
  title = {Synthesizing Text-to-{SQL} Data from Weak and Strong {LLM}s},
  booktitle = {Proceedings of the 62nd Annual Meeting of the Association for Computational Linguistics (Volume 1: Long Papers)},
  series = {ACL~'24},
  pages = {7864--7875},
  address = {Bangkok, Thailand},
  publisher = {Association for Computational Linguistics},
  year = {2024},
  doi = {10.18653/v1/2024.acl-long.425},
  url = {https://aclanthology.org/2024.acl-long.425/},
}

@inproceedings{song2025jolt,
  author = {Song, Jinwang and Zan, Hongying and Zhang, Kunli and Mu, Lingling and Han, Yingjie and Hua, Haobo and Peng, Min},
  title = {{JOLT-SQL}: Joint Loss Tuning of Text-to-{SQL} with Confusion-aware Noisy Schema Sampling},
  booktitle = {Proceedings of the 2025 Conference on Empirical Methods in Natural Language Processing},
  series = {EMNLP~'25},
  address = {Suzhou, China},
  publisher = {Association for Computational Linguistics},
  pages = {6040--6053},
  year = {2025},
  doi = {10.18653/v1/2025.emnlp-main.308},
  url = {https://aclanthology.org/2025.emnlp-main.308/},
}

@inproceedings{park2024grammaraligned,
  author = {Park, Kanghee and Wang, Jiayu and Berg-Kirkpatrick, Taylor and Polikarpova, Nadia and D'Antoni, Loris},
  title = {Grammar-Aligned Decoding},
  booktitle = {Advances in Neural Information Processing Systems},
  series = {NeurIPS~'24},
  volume = {37},
  pages = {24547--24568},
  address = {Vancouver, BC, Canada},
  publisher = {Curran Associates, Inc.},
  year = {2024},
  doi = {10.52202/079017-0774},
  url = {https://proceedings.neurips.cc/paper_files/paper/2024/hash/2bdc2267c3d7d01523e2e17ac0a754f3-Abstract-Conference.html},
}

@inproceedings{ma2025sqlr1,
  author = {Ma, Peixian and Zhuang, Xialie and Xu, Chengjin and Jiang, Xuhui and Chen, Ran and Guo, Jian},
  title = {{SQL-R1}: Training Natural Language to {SQL} Reasoning Model By Reinforcement Learning},
  booktitle = {Advances in Neural Information Processing Systems},
  series = {NeurIPS~'25},
  volume = {38},
  pages = {174505--174537},
  address = {San Diego, CA, USA},
  publisher = {Curran Associates, Inc.},
  year = {2025},
  doi = {10.52202/085713-5805},
  url = {https://proceedings.neurips.cc/paper_files/paper/2025/hash/ff00aae6ce2b5e519ca4cf769f0c4a47-Abstract-Conference.html},
}

@inproceedings{pourreza2025chase,
  author = {Pourreza, Mohammadreza and Li, Hailong and Sun, Ruoxi and Chung, Yeounoh and Talaei, Shayan and Kakkar, Gaurav Tarlok and Gan, Yu and Saberi, Amin and {\"O}zcan, Fatma and Ar{\i}k, Sercan {\"O}.},
  title = {{CHASE-SQL}: Multi-Path Reasoning and Preference Optimized Candidate Selection in Text-to-{SQL}},
  booktitle = {Proceedings of the Thirteenth International Conference on Learning Representations},
  series = {ICLR~'25},
  address = {Singapore},
  publisher = {OpenReview.net},
  year = {2025},
  url = {https://proceedings.iclr.cc/paper_files/paper/2025/hash/974ff7b5bf08dbf9400b5d599a39c77f-Abstract-Conference.html},
}
\clearpage
\appendix
\lstdefinestyle{appendixcode}{
    basicstyle=\ttfamily\fontsize{9}{10}\selectfont,
    breaklines=true,
    breakatwhitespace=false,
    columns=fullflexible,
    keepspaces=true,
    frame=single,
    rulecolor=\color{black!20},
    backgroundcolor=\color{gray!5},
    xleftmargin=2pt,
    xrightmargin=2pt,
    aboveskip=4pt,
    belowskip=4pt
}
\lstset{style=appendixcode}
\section{Structural Measure Details}
\label{app:measures}
\paragraph{Parse-failure conventions.}
Our evaluation implementation assigns zero to counts, majority ratios, and entropies when the selected structural subset is empty. Dataset summaries average these values over all questions. The denominator therefore includes questions with no usable structure. An average below one can consequently reflect structural coverage as well as concentration among available structures.

\paragraph{Subset interpretation.}
The metric definitions and abbreviations in Section~\ref{Structural-Metrics} apply throughout. CA and $D_c$ use parsed execution-correct outputs, while D uses all parsed outputs. A low CA can reflect either \problem or an empty correct structural subset. LAR inherits this coverage dependence; HDR explicitly requires multiple parsed correct structures and therefore identifies \problem at high accuracy.

\section{Additional Scope-Validation Statistics}
\label{app:scope}
Table~\ref{tab:scope-extra} complements the main-text subset results with parsing coverage and structural diversity. Each row summarizes 100 questions with ten generations per question; the zero convention above applies to all structural averages.
\begin{table*}[t]
\centering
\small
\begin{tabularx}{\textwidth}{ll*{5}{>{\centering\arraybackslash}X}}
\toprule
\rowcolor{tableHeader!65!white}
Subset & Model & PR$\uparrow$ & D & $D_c$ & CA$\uparrow$ & $H_c$\\
\midrule
\rowcolor{modelGPT} BIRD Mini-Dev & \modelentry{gpt5mini} & 0.657 & 3.380 & 0.560 & 0.091 & 0.200\\
\rowcolor{modelClaude}  & \modelentry{claudesonnet} & 0.496 & 1.060 & 0.290 & 0.210 & 0.041\\
\rowcolor{modelDeepSeek}  & \modelentry{deepseek} & 0.399 & 1.600 & 0.500 & 0.212 & 0.152\\
\rowcolor{modelGPT} Spider-Syn & \modelentry{gpt5mini} & 0.713 & 2.630 & 0.540 & 0.186 & 0.177\\
\rowcolor{modelClaude}  & \modelentry{claudesonnet} & 0.686 & 1.200 & 0.410 & 0.410 & 0.000\\
\rowcolor{modelDeepSeek}  & \modelentry{deepseek} & 0.453 & 1.260 & 0.530 & 0.300 & 0.126\\
\rowcolor{modelGPT} Dr.Spider & \modelentry{gpt5mini} & 0.668 & 2.610 & 0.300 & 0.121 & 0.092\\
\rowcolor{modelClaude}  & \modelentry{claudesonnet} & 0.720 & 1.460 & 0.290 & 0.276 & 0.010\\
\rowcolor{modelDeepSeek}  & \modelentry{deepseek} & 0.470 & 1.490 & 0.420 & 0.287 & 0.072\\
\bottomrule
\end{tabularx}
\caption{Parsing coverage varies substantially across the additional subsets. \modelname{gpt5mini} has the greatest overall structural diversity in each subset, while \modelname{claudesonnet} has no observed correct-output diversity on Spider-Syn.}
\label{tab:scope-extra}
\end{table*}

\section{Additional Error Analysis}
\label{appendix:error}

To better understand the behavioral differences between traditional
direct SQL generation and the proposed compile-style generation
paradigm, we conduct an additional qualitative error analysis on
representative examples from the Spider development set.

\subsection{Diagnostic Categories}

We assign inspected queries to six diagnostic categories. Five describe errors, and the sixth records structural variation among execution-correct alternatives:

\begin{itemize}

\item \textbf{Schema linking.}
Incorrect or missing references to tables, columns, or foreign-key
relationships between tables.

\item \textbf{Logical form.}
Errors in high-level query logic such as aggregation, grouping,
nested queries, or set operations.

\item \textbf{Join path.}
Incorrect or incomplete join paths between tables, including missing
joins or incorrect join conditions.

\item \textbf{Predicate.}
Incorrect or missing filtering conditions in \texttt{WHERE} or
\texttt{HAVING} clauses, such as incorrect comparison operators or
incorrect attribute references.

\item \textbf{Structural variation.}
Queries that produce correct execution results but differ substantially
in structure from the gold query or other correct solutions.

\item \textbf{Compilation.}
Errors introduced by the structured intermediate representation or
the compilation process in the compile-style generation pipeline.

\end{itemize}

Schema linking is also an explicit target of recent methods: JOLT-SQL jointly optimizes schema linking and SQL generation and uses confusion-aware noisy schema sampling to improve robustness~\citep{song2025jolt}. Our diagnostic categories distinguish these schema-reference errors from errors in query logic and from variation among execution-correct queries.

\subsection{Category Distribution}

Table~\ref{tab:error_distribution} compares category proportions for Direct SQL and compile-style generation. Each ratio represents the fraction of inspected examples assigned to the corresponding primary category. We inspected 100 examples for each setting, sampled from cases with execution errors or structural disagreement. One author annotated the examples and a second author checked the labels. We assigned mutually exclusive primary categories, choosing the earliest issue that determined failure or structural divergence when multiple issues occurred. Here, $\Delta$ is the compile-style proportion minus the Direct SQL proportion. The proportions summarize these selected cases of execution errors and structural disagreement.

\begin{table}[t]
\centering
\small
\resizebox{\columnwidth}{!}{
\begin{tabular}{l>{\columncolor{tableLilac!65!white}}r>{\columncolor{tableCyan!40!white}}r r}
\toprule
\rowcolor{tableHeader!65!white}
Category & Direct SQL & Compile-style & $\Delta$ \\
\midrule
Schema linking & 0.20 & 0.13 & -0.07 \\
Logical form & 0.18 & 0.15 & -0.03 \\
Join path & 0.14 & 0.09 & -0.05 \\
Predicate & 0.16 & 0.10 & -0.06 \\
Structural variation & 0.28 & 0.40 & +0.12 \\
Compilation & 0.04 & 0.13 & +0.09 \\
\bottomrule
\end{tabular}
}
\caption{The inspected compile-style sample has smaller shares of schema, logic, join, and predicate errors, but larger shares of structural variation and compilation errors.}
\label{tab:error_distribution}
\end{table}

As shown in Table~\ref{tab:error_distribution}, the selected compile-style sample has smaller proportions of several semantic-error categories, including schema linking,
logical form mistakes, incorrect join paths, and predicate errors.
However, two categories increase: structurally diverse but executable
queries (structural variation) and errors in the intermediate representation or its compilation (compilation).
This pattern is consistent with the quantitative observations
reported in Experiment~3.

\subsection{Representative Case Studies}

We further illustrate typical failure modes using representative
examples from the Spider development set.

\paragraph{Direct SQL Execution Failures.}

The inspected examples include schema-linking and join-path errors in direct generation. The intermediate representation makes table relations explicit, allowing us to inspect the model's table choices and joins before compilation.

For example, for the question:

\begin{samepage}\begin{quote}
``Which model of the car has the minimum horsepower?''
\par\end{quote}\end{samepage}

the gold query joins \texttt{CAR\_NAMES} and \texttt{CARS\_DATA}
to retrieve the corresponding model. An inspected baseline output
omits the join; such queries directly access the
\texttt{CARS\_DATA} table, resulting in incorrect results. In contrast,
compile-style generation constructs the join structure explicitly
in the intermediate representation, producing SQL queries that are
structurally equivalent to the gold query.

\paragraph{Examples of \problem.}

We also observe \problem in the inspected baseline outputs: multiple generated queries return the correct result but have different canonical structures. For example, for the question:

\begin{samepage}\begin{quote}
``Show all countries and the number of singers in each country.''
\par\end{quote}\end{samepage}

the inspected baseline outputs include structurally distinct SQL
queries, including variants with redundant subqueries or unnecessary
\texttt{DISTINCT} operators. Although these queries produce identical
results, they correspond to different AST structures and therefore
increase structural entropy. The compile-style example uses a simple
\texttt{GROUP BY} aggregation.

\paragraph{Paraphrase Sensitivity.}

Semantically equivalent questions can trigger different query
structures in baseline generation. For example, the two questions

\begin{samepage}\begin{quote}
``Show all countries and the number of singers in each country'' \\
``How many singers are from each country?''
\par\end{quote}\end{samepage}

share the same semantic intent and annotated reference SQL query. However, the inspected baseline
generations use different structural patterns for the two
inputs, including variations involving subqueries or derived tables.
This example illustrates why agreement across paraphrases is evaluated separately from agreement across repeated generations of one input.

\paragraph{Compile-style Failures.}

Despite the overall improvements, compile-style generation introduces
new failure modes related to the intermediate representation and
compilation process. For instance, when answering the question

\begin{samepage}\begin{quote}
``How many car models are produced in the USA?''
\par\end{quote}\end{samepage}

the correct query requires joining \texttt{MODEL\_LIST},
\texttt{CAR\_MAKERS}, and \texttt{COUNTRIES}. In some compile-style
generations, the intermediate representation omits the final join to
\texttt{COUNTRIES} and instead applies the filtering condition directly
to the \texttt{CAR\_MAKERS} table. This leads to execution errors despite
the structured generation pipeline.

These cases illustrate successful query construction as well as failures in the intermediate representation. They highlight specific joins and predicates to examine when refining the representation and compiler.

\section{Implementation Details}
\label{app:implementation}

This section describes the implementation details of \method, including SQL parsing and canonicalization, canonical AST representation, execution evaluation, and generation settings.

\subsection{SQL Parsing and Canonicalization}

We parse all generated SQL queries using the \texttt{sqlglot} parser, which provides robust SQL parsing and AST construction across multiple SQL dialects. In our experiments we use the SQLite dialect to ensure compatibility with the Spider benchmark databases.

To remove surface-level differences that are irrelevant to program structure, we apply both text-level normalization and AST-level canonicalization.

\paragraph{Text-level normalization.}
After rendering the parsed AST back into SQL text, we perform lightweight normalization including: (1) removing trailing semicolons and leading/trailing whitespace, (2) replacing line breaks with spaces and collapsing redundant whitespace, (3) removing simple column aliases (e.g., \texttt{SELECT col AS x}), and (4) converting the SQL string to lowercase.

\paragraph{AST-level canonicalization.}
To ensure that structurally equivalent queries share the same representation, we further canonicalize the AST structure:

\begin{itemize}
\item \textbf{Alias normalization.} Table aliases are renamed according to their first appearance order (e.g., \texttt{t1, t2, ...}). All column references are updated accordingly so that structurally identical joins remain equivalent under different alias choices.

\item \textbf{Logical operator normalization.} Nested \texttt{AND} predicates are flattened into condition lists, sorted by their SQL string representation, and reconstructed into canonical AST nodes. This ensures that logically equivalent expressions such as \texttt{A AND B} and \texttt{B AND A} share the same canonical form.
\end{itemize}

These normalization steps are shared across all experiments (Exp1--Exp4) and applied to both baseline and compile-style generated queries.

\subsection{Canonical AST Representation}

For each SQL query we obtain a canonical representation through the following pipeline:

\begin{enumerate}
\item Parse the SQL query into an AST using \texttt{sqlglot};
\item Apply AST-level canonicalization (alias normalization and logical operator normalization);
\item Render the canonical AST back into SQL;
\item Apply text-level normalization.
\end{enumerate}

The resulting canonical SQL string serves as the canonical comparison representation and is used as the basis for all structural measures in our analysis.

For each question with $k$ generated SQL queries, we compute the frequency distribution over canonical structures. Structural statistics are then derived from this distribution:

\begin{itemize}
\item \textbf{Distinct structures:} the number of unique canonical SQL representations.
\item \textbf{Majority structure ratio:} the relative frequency of the most common structure among the $k$ generations.
\item \textbf{Structural entropy:} the Shannon entropy of the structure distribution, measuring structural diversity.
\end{itemize}

To compare generated structures with annotated reference queries, the annotated reference SQL is canonicalized using the same pipeline, and structural matches are determined by canonical representation equality.

\subsection{Execution Environment}

The main experiments use the Spider SQLite database files; the additional scope checks use the corresponding benchmark databases. We compare query execution results against the annotated reference result on the same database instance.

For each generated SQL query, we execute the query against the corresponding SQLite database and obtain the result set. A generated query is considered execution-correct if its execution result exactly matches the result produced by the annotated reference SQL query. Queries that produce execution errors or mismatched results are considered incorrect.

Execution feedback also serves as a learning signal in recent Text-to-SQL systems: SQL-R1 uses rewards for successful execution and result correctness~\citep{ma2025sqlr1}. Structural evaluation can complement such feedback by distinguishing query constructions within the execution-correct subset.

\subsection{Generation Settings}

For each question we generate multiple candidate SQL queries from each model to analyze structural variation.

\paragraph{Sampling configuration.}
We sample $k=10$ SQL queries for every $(question, schema)$ pair. For APIs that return one sample per request, we make independent calls and aggregate the outputs.

\paragraph{Decoding parameters.}
Unless restricted by the model API, we use stochastic decoding with temperature $=1.0$ and default nucleus sampling parameters. Stochastic decoding lets us analyze structural variation across repeated generations.

Decoding constraints can also affect output distributions. \citet{park2024grammaraligned} show that conventional grammar-constrained decoding can distort the model distribution. Structural diversity should therefore be interpreted together with the decoding policy used to produce the samples.

\paragraph{Compile-style generation.}
In Experiment~3, the compile-style setting asks models to generate a structured intermediate representation for compilation into SQL. The structured representation is expressed as JSON and contains fields such as \texttt{select}, \texttt{from}, \texttt{joins}, \texttt{where}, \texttt{group\_by}, \texttt{having}, \texttt{order\_by}, and \texttt{limit}. A deterministic compiler converts the JSON representation into executable SQL. The compiled SQL is then evaluated using the same canonicalization and execution evaluation pipeline as the baseline generation.

\section{Prompt and Generation Settings}
\label{app:prompts}

This section provides the prompt templates used in our experiments.

\subsection{Prompt for Direct SQL Generation}

In the baseline setting, models are prompted to directly generate a SQL query 
given a natural language question and the corresponding database schema.

\begin{lstlisting}[style=appendixcode]
You are an expert SQL generator.
Use SQLite dialect.
Only output ONE SQL query, no explanation.

Database ID: <db_id>

Database schema (JSON):
<schema_json>

Question:
<natural_language_question>

SQL:
\end{lstlisting}

\subsection{Paraphrase Generation and Checking}
\label{app:paraphrases}
Candidate paraphrases were generated using an LLM prompt and manually checked before use. The prompt is:
\begin{lstlisting}
Rewrite the following Text-to-SQL question into semantically equivalent paraphrases. Preserve the requested attributes, aggregation, filters, comparison operators, ordering, limits, and all schema-specific meanings.
Do not add or remove constraints. Do not make the question more ambiguous.

Original question:
<question>

Output <n> paraphrases, one per line.
\end{lstlisting}

We excluded paraphrases that changed selected columns, aggregation targets, predicates, comparison relations, ordering, limits, or intended schema entities. We also excluded variants that added conditions, omitted necessary constraints, or introduced ambiguity. These criteria guide the check that each paraphrase preserves the original intent.

\subsection{Prompt for Compile-style JSON Generation}

In the compile-style setting, the model is instructed to generate a structured
JSON representation of the SQL query for compilation into SQL.
The prompt used for this setting is shown below.

\begin{lstlisting}[style=appendixcode]
You are an expert Text-to-SQL system for the Spider benchmark.
Your task is to write a structured JSON representation of a SQL query
for the given question and database schema.

Requirements:

- Use ONLY tables and columns that exist in the provided schema.
- Assume the database uses the SQLite dialect.
- You MUST output a single JSON object, and nothing else (no explanations).
- The JSON must describe the logical structure of the SQL query with the following fields:
  - type: "query"
  - query: {
    select: [ ... ],
    from: { ... },
    joins: [ ... ],
    where: [ ... ],
    group_by: [ ... ],
    having: [ ... ],
    order_by: [ ... ],
    limit: ...,
    distinct: ...
  }
- Do NOT include any natural language text in the JSON.

Database ID: {db_id}

Database schema (JSON):
{schema_json}

Question:
{question}

Now output ONLY the JSON object for the query structure:
\end{lstlisting}

\section{Additional Structural Examples}
\label{app:examples}

To complement the quantitative structural measures reported in the main experiments, we present several representative examples illustrating common forms of structural variation in LLM-generated SQL queries.

\paragraph{Example 1: Execution-correct but structurally different queries}
\label{appendixexample1}
\textbf{Question.}  
Which model of the car has the minimum horsepower? (\texttt{db\_id = car\_1})

\Needspace{82pt}
\textbf{Gold SQL.}

\begin{lstlisting}[style=appendixcode]
SELECT T1.Model
FROM CAR_NAMES AS T1
JOIN CARS_DATA AS T2 ON T1.MakeId = T2.Id
ORDER BY T2.Horsepower ASC
LIMIT 1;
\end{lstlisting}

\Needspace{92pt}
\textbf{Generated SQL (baseline).}

\begin{lstlisting}[style=appendixcode]
SELECT DISTINCT cn.Model
FROM cars_data cd
JOIN car_names cn ON cd.Id = cn.MakeId
WHERE cd.Horsepower = (
  SELECT MIN(Horsepower) FROM cars_data
);
\end{lstlisting}

\begin{figure*}[t]
    \centering
    \begin{subfigure}[t]{0.48\textwidth}
        \centering
        \includegraphics[width=0.93\linewidth]{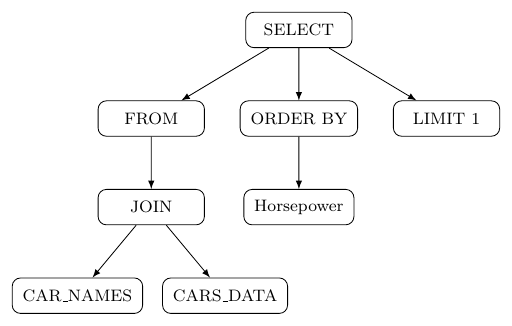}
        \caption{AST of gold SQL query}
    \end{subfigure}
    \hfill
    \begin{subfigure}[t]{0.48\textwidth}
        \centering
        \includegraphics[width=\linewidth]{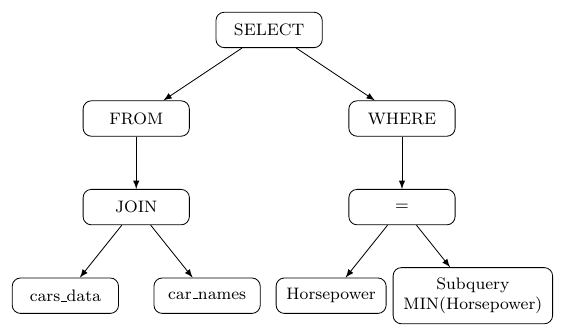}
        \caption{AST of generated SQL query}
    \end{subfigure}

    \caption{AST comparison for the minimum-horsepower example in Appendix~\ref{appendixexample1}. The reference uses ordering and a row limit; the generated query uses an aggregate subquery. Both return the same result on the evaluated database.}
    \label{fig:ast_comparison_example1}
\end{figure*}

Although both queries return identical execution results, their canonical AST structures differ substantially, as illustrated in Figure~\ref{fig:ast_comparison_example1}.
The gold query computes the minimum value using an \texttt{ORDER BY + LIMIT} structure, while the generated query uses an aggregate subquery with \texttt{MIN(Horsepower)} and an equality predicate. This introduces a nested subquery node in the AST, resulting in a different structural representation.

\paragraph{Example 2: \problem across paraphrases}

Consider the following pair of paraphrased questions from the Spider dataset (\texttt{db\_id = concert\_singer}):

\textbf{Question A.}  
Show all countries and the number of singers in each country.

\textbf{Question B.}  
How many singers are from each country?

Both questions correspond to the same annotated reference SQL query:

\begin{lstlisting}[style=appendixcode]
SELECT country, COUNT(*)
FROM singer
GROUP BY country;
\end{lstlisting}

However, the baseline model produces slightly different structural variants:

\begin{lstlisting}[style=appendixcode]
-- Variant A
SELECT Country, COUNT(Singer_ID)
FROM singer
GROUP BY Country;

-- Variant B
SELECT Country, COUNT(*)
FROM singer
GROUP BY Country;
\end{lstlisting}

These generated queries illustrate \problem across equivalent phrasings: both return the correct result on the evaluated database, but their ASTs differ in the aggregation argument (\texttt{COUNT(*)} vs.\ \texttt{COUNT(Singer\_ID)}).

\paragraph{Example 3: Compile-style correcting structural errors}

\textbf{Question.}  
What are the names and ids of all countries with at least one car maker? (\texttt{db\_id = car\_1})

\Needspace{92pt}
\textbf{Gold SQL.}

\begin{lstlisting}[style=appendixcode]
SELECT T1.CountryName, T1.CountryId
FROM COUNTRIES AS T1
JOIN CAR_MAKERS AS T2 ON T1.CountryId = T2.Country
GROUP BY T1.CountryId
HAVING COUNT(*) >= 1;
\end{lstlisting}

\Needspace{72pt}
\textbf{Baseline SQL (typical failure).}

\begin{lstlisting}[style=appendixcode]
SELECT CountryName, CountryId
FROM COUNTRIES
JOIN CAR_MAKERS
ON COUNTRIES.CountryId = CAR_MAKERS.Country;
\end{lstlisting}

The illustrated baseline output omits the \texttt{GROUP BY} and \texttt{HAVING} clauses, producing incorrect query results.

\Needspace{102pt}
\textbf{Compile-style SQL.}

\begin{lstlisting}[style=appendixcode]
SELECT COUNTRIES.CountryName, COUNTRIES.CountryId
FROM COUNTRIES
JOIN CAR_MAKERS
ON COUNTRIES.CountryId = CAR_MAKERS.Country
GROUP BY COUNTRIES.CountryId
HAVING COUNT(*) >= 1;
\end{lstlisting}

In the compile-style generation setting, the structured intermediate representation explicitly models aggregation components such as \texttt{group\_by} and \texttt{having}. The displayed compiled SQL contains the aggregation structure and closely matches the gold AST.

\section{Example of the Compile-style Generation Pipeline}
\label{app:pipeline}

To illustrate the compile-style generation process used in Experiment~3, 
we present a concrete example showing the full pipeline from natural language 
to structured intermediate representation, internal AST, and final SQL query.

\subsection{Input Natural Language Query}

\textbf{Question.}

\begin{samepage}\begin{quote}
Show the name and capacity of the stadium with the largest average attendance.
\par\end{quote}\end{samepage}

\textbf{Database schema (simplified).}

\begin{lstlisting}[style=appendixcode]
{
  "tables": [
    {"table_name": "stadium",
     "columns": ["Stadium_ID", "Name", "Location", "Capacity", "Average"]},
    {"table_name": "concert",
     "columns": ["concert_ID", "Stadium_ID", "Year"]}
  ],
  "foreign_keys": [
    {"source_table": "concert", "source_column": "Stadium_ID",
     "target_table": "stadium", "target_column": "Stadium_ID"}
  ]
}
\end{lstlisting}

\subsection{LLM Output: Structured JSON Intermediate Representation}

In compile-style generation, the model generates a structured JSON representation that encodes the query components for compilation into SQL.

The two entries in \texttt{select} preserve the requested output attributes, \texttt{Name} and \texttt{Capacity}. The ranking attribute, \texttt{Average}, appears only in \texttt{order\_by}; it need not be returned as an output column. Descending order and a limit of one implement the superlative in the question. Here, ``average attendance'' refers to the stored \texttt{Average} column, so the representation contains no aggregation or grouping operation.

Although the simplified schema also includes \texttt{concert}, every required attribute belongs to \texttt{stadium}. The empty \texttt{joins} list therefore makes the single-table choice explicit. The empty predicate lists indicate that the question requests no additional filtering.

\begin{lstlisting}[style=appendixcode]
{
  "type": "query",
  "query": {
    "select": [
      {"expr": {"col": ["stadium", "Name"]}, "alias": null},
      {"expr": {"col": ["stadium", "Capacity"]}, "alias": null}
    ],
    "from": {"table": "stadium", "alias": null},
    "joins": [], "where": [],
    "group_by": [], "having": [],
    "order_by": [
      {"expr": {"col": ["stadium", "Average"]}, "direction": "desc"}
    ],
    "limit": 1, "distinct": false
  }
}
\end{lstlisting}

This JSON structure explicitly represents the logical components of the query,
including \texttt{select}, \texttt{from}, \texttt{order\_by}, and \texttt{limit}.

\subsection{Intermediate AST Representation}

The compiler then converts the JSON structure into an internal AST:

\begin{lstlisting}[style=appendixcode]
Select(
  columns = [
    Column(table="stadium", name="Name"),
    Column(table="stadium", name="Capacity")
  ],
  from = Table(name="stadium"),
  joins = [], where = [],
  group_by = [], having = [],
  order_by = [
    OrderBy(
      expr = Column(table="stadium", name="Average"),
      direction = DESC
    )
  ],
  limit = 1)
\end{lstlisting}

This AST provides a canonical structural representation of the query,
which can be analyzed and compared across different generations.

\subsection{Compiled SQL Query}

Finally, the AST is deterministically compiled into an executable SQL query:

\begin{lstlisting}[style=appendixcode]
SELECT stadium.Name, stadium.Capacity
FROM stadium
ORDER BY stadium.Average DESC
LIMIT 1;
\end{lstlisting}

This compiled SQL query is then used for execution evaluation and 
structural analysis in our experiments.

\subsection{Reading and Checking the Pipeline}
\label{app:pipeline-checks}

The example separates decisions made during generation from their realization in SQL. The model chooses the table, output attributes, ordering expression, and row limit in the JSON representation. The compiler carries these choices into the displayed AST and SQL. A deterministic translation makes a fixed representation reproducible, but a wrong table or ordering choice in that representation can still yield an incorrect query.

\paragraph{Component correspondence.}
Each populated field has a visible counterpart. The selected expressions become \texttt{Column} nodes and the SQL projection; the source becomes the \texttt{Table} node and \texttt{FROM} clause. The ordering field becomes an \texttt{OrderBy} node and \texttt{ORDER BY stadium.Average DESC}, while the row limit remains one. Empty lists introduce no join, predicate, or grouping clauses. This correspondence supports inspection beyond the execution result.

\paragraph{Semantic boundary cases.}
The query returns at most one row. If several stadiums share the largest average attendance, the displayed ordering contains no secondary key to select a particular stadium among them. A formulation that filters on the maximum value can instead return multiple rows in that case. Thus, agreement on the evaluated database does not establish equivalence under changes to the data. Testing tied values would help distinguish these constructions; no such additional experiment is reported here.

\paragraph{Connection to structural evaluation.}
The displayed internal AST explains compilation. Structural scoring subsequently applies the shared SQL parsing and canonicalization procedure to the compiled SQL, as described in Appendix~\ref{app:implementation}. Direct and compiled SQL thus enter the same evaluator; JSON formatting differences are not counted as SQL structures.

\end{document}